\documentclass[10pt,twocolumn,letterpaper]{article}

\usepackage{wacv}
\usepackage{times}
\usepackage{epsfig}
\usepackage{graphicx}
\usepackage{amsmath}
\usepackage{amssymb}

\usepackage{adjustbox}
\usepackage{subcaption}
\usepackage{enumitem}

\usepackage{algorithm}
\usepackage{algorithmic}

\usepackage{flushend}

\usepackage[colorlinks,pagebackref=true,citecolor=green,bookmarks=false,hypertexnames=true]{hyperref}
\usepackage{nopageno}

\def\wacvPaperID{0259} 

\wacvfinalcopy 

\ifwacvfinal
\def\assignedStartPage{1} 
\fi

\def\##1{\relax\ifmmode\mathchoice
{\mbox{\boldmath$\displaystyle#1$}}
{\mbox{\boldmath$\textstyle#1$}}
{\mbox{\boldmath$\scriptstyle#1$}}
{\mbox{\boldmath$\scriptscriptstyle#1$}}\else
\hbox{\boldmath$\textstyle$}\fi}

\ifwacvfinal
\else
\fi

\begin{document}

\title{Let's Get Dirty: GAN Based Data Augmentation for \\Camera Lens Soiling Detection in Autonomous Driving}

\author{%
Michal U\v{r}i\v{c}\'{a}\v{r}$^{1}\thanks{Most of this work was done while Michal U\v{r}i\v{c}\'{a}\v{r} and Pavel K\v{r}\'{i}\v{z}ek were employed by Valeo.}$ , 
Ganesh Sistu$^{2}$, 
Hazem Rashed$^{2}$, 
Anton\'{i}n Vobeck\'{y}$^{3,2}$, \\
Varun Ravi Kumar$^{2}$,
Pavel K\v{r}\'{i}\v{z}ek$^{1*}$, 
Fabian B\"{u}rger$^{2}$ 
and Senthil Yogamani$^{2}$ \\ 
{\normalsize 
$^{1}$Independent Researcher \hspace{0.3cm} 
$^{2}$Valeo \hspace{0.3cm} 
$^{3}$CTU in Prague \hspace{0.3cm} 
} \\
{
\tt \small
uricar.michal@gmail.com \hspace{0.3cm}
antonin.vobecky@cvut.cz \hspace{0.3cm}
firstname.lastname@valeo.com}
}

\maketitle


\begin{abstract}

Wide-angle fisheye cameras are commonly used in automated driving for parking and low-speed navigation tasks. Four of such cameras form a surround-view system that provides a complete and detailed view of the vehicle. These cameras are directly exposed to harsh environmental settings and can get soiled very easily by mud, dust, water, frost. Soiling on the camera lens can severely degrade the visual perception algorithms, and a camera cleaning system triggered by a soiling detection algorithm is increasingly being deployed. While adverse weather conditions, such as rain, are getting attention recently, there is only limited work on general soiling. The main reason is the difficulty in collecting a diverse dataset as it is a relatively rare event. 

We propose a novel GAN based algorithm for generating unseen patterns of soiled images. Additionally, the proposed method automatically provides the corresponding soiling masks eliminating the manual annotation cost. Augmentation of the generated soiled images for training improves the accuracy of soiling detection tasks significantly by $18\%$ demonstrating its usefulness. The manually annotated soiling dataset and the generated augmentation dataset will be made public. We demonstrate the generalization of our fisheye trained GAN model on the Cityscapes dataset. We provide an empirical evaluation of the degradation of the semantic segmentation algorithm with the soiled data. 

\end{abstract}



\section{Introduction} 
\label{sec:intro}

\begin{figure}[tb]
    \centering
    \includegraphics[width=0.8\linewidth]{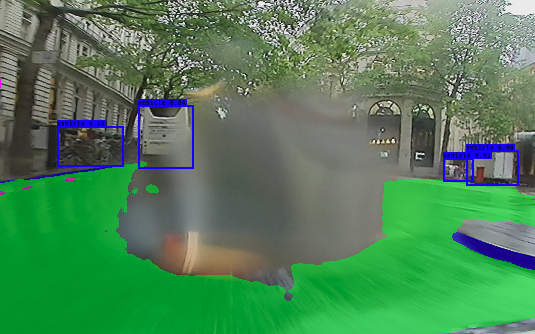}
    \caption{The example of a semi-transparent soiling in form of a water drop on the camera lens. The detection of the bus behind the water drop works still well, while the road segmentation (green) is highly degraded in the soiled region. In this scenario, a soiling detection algorithm is used to trigger a camera cleaning system which restores the lens hardware.}
    \label{fig:soiling_example_drop}
    \vspace{-6mm}
\end{figure}
Level $5$ autonomous driving (\cite{SAE_automation}) stands out as a challenging goal of a large part of the computer vision and machine learning community. Due to this problem's difficulty, a combination of various sensors is necessary to build a safe and robust system. Even just considering cameras, a combination of a narrow field of view (FOV) long-range sensing and wide FOV short-range sensing is necessary. The latter is achieved by fisheye cameras, which provide complete near-field sensing, up to $10$m around the vehicle. One use-case where it is necessary is the fish-bone parking~\cite{heimberger2017computer}, where ultrasonic sensors and narrow FOV cameras are not sufficient. Without exaggeration, these fisheye cameras are becoming \textit{de facto} standard in low-speed maneuvering applications, like parking and traffic jams assist~\cite{maddu2019fisheyemultinet}.
\begin{figure*}[tb]
    \centering
    \includegraphics[width=0.8\linewidth]{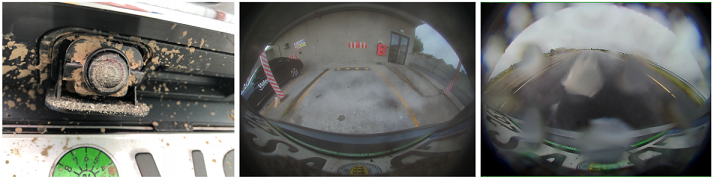}
    \caption{Automotive surround view cameras are exposed to harsh environmental setup. Left: camera lens covered by mud. Middle: image produced by the soiled camera from the left picture. Right: camera lens soiled during heavy rain.}
    \label{fig:soiling_setup}
\end{figure*}
Enormous progress can be noticed in typical image processing tasks, such as semantic segmentation or object detection~\cite{siam2017deep, siam2018rtseg, siam2018modnet}, which is mainly attributed to the prevailing success of deep learning. However, there are other less ``popular'' problems slowly getting into attention, which have to be solved as well for the ultimate goal of the full Level $5$ autonomy. 

One of these problems is the reliability of the sensory signal, which in the case of surround-view cameras means, \textit{inter alia}, the ability to detect soiling on the camera lens. Failure to recognize severe weather conditions leading to a deterioration of the image quality to such a level that any further image processing is unreliable. Figure~\ref{fig:soiling_setup} shows how the surround-view camera can get soiled and the corresponding image output, as well as an example of images taken during a~heavy rain. It usually happens when the tire is splashing mud or water from the road or due to wind depositing dust on the lens. 
Figure~\ref{fig:soiling_example_drop} shows an example of the strong impact of a significant water drop on the camera lens for object detection and semantic segmentation tasks.

In this work, we focus on soiling caused by a variety of unwanted particles reposing on the camera lens. The source of these particles is mostly mud, dirt, water, or foam created by a detergent. Based on the state of aggregation, such soiling can be either static (e.g., highly viscous mud tends to dry up very quickly, so it does not change its position on the output image over time) or dynamic (mostly water, and foam). Because of that, the acquisition of suitable data for training machine learning models or merely testing the effect on existing classification models is quite tedious. Human annotation is very time demanding, costly, and not very reliable since the precise labeling of soiling on a single image can sometimes be very challenging, e.g., due to the unclear boundary. We want to emphasize that the soiling detection task is necessary for an autonomous driving system as it is used to trigger a camera cleaning system that restores the lens~\cite{uricar2019soilingnet}. It is complementary to building segmentation or object detection models robust to soiling without an explicit soiling detection step. Our contributions include:
\begin{itemize}[nolistsep]
    \item A proposition of a baseline pipeline for an opaque soiling generation, based on CycleGAN~\cite{Zhu-ICCV-2017} and soiling segmentation learned from weak labels. 
    \item A novel DirtyGAN network, which is an end-to-end generalization of the baseline pipeline. 
    \item Public release of an artificial soiling dataset as a companion to the recently published \textit{WoodScape Dataset}~\cite{woodscape-2019}, coined \textit{Dirty WoodScape Dataset}, to encourage further research in this area. 
    \item An empirical evaluation of the degradation of semantic segmentation algorithms on soiled images. 
\end{itemize}

The rest of the paper is organized as follows. Section~\ref{sec:related} covers the related work and contrasts the soiling scenarios with adverse weather conditions. In Section~\ref{sec:proposed}, we give a detailed description of the proposed algorithms. Section~\ref{sec:experiments} provides an evaluation of the quality of the generated images and quantifies the improvement of the soiling detection algorithm by adding generated images to training. Also, it describes the empirical evaluation of the degradation of semantic segmentation in the presence of soiling. Finally, Section~\ref{sec:conclusions} concludes the paper.


\section{Related Work} 
\label{sec:related}

As the visual perception modules for autonomous driving are becoming more mature, there is much recent effort to make them more robust to adverse weather conditions. It can be seen by the popular CVPR workshop ``Vision for all seasons''\footnote{\url{https://vision4allseasons.net/}}, which focuses on the performance of computer vision algorithms during adverse weather conditions for autonomous driving. However, there is very little work on the related but different lens soiling problem. The two problems are similar in how they degrade image quality and can severely affect visual perception performance. Yet there are substantial differences. 

The first significant difference is that soiling on the lens can be removed by a camera cleaning system that either sprays water or uses a more sophisticated ultrasonic hardware~\cite{uricar2019soilingnet}. Secondly, there is temporal consistency for soiling where mud or water droplets remain static typically or sometimes have low-frequency dynamics of moving water droplets in contrast to higher variability in adverse weather scenes. Thus this temporal structure can be exploited further for soiling scenarios. Finally, soiling can cause more severe degradation as opaque mud soiling can completely block the camera.

We focus on the generic soiling detection task in this paper. Even disregarding camera cleaning, soiling detection is still needed to increase the uncertainty of vision algorithms in the degraded areas. The task of soiling detection on camera lenses in autonomous driving is shortly described in~\cite{Uricar-2019a}. The authors present a sort of a proof of concept idea on how Generative Adversarial Networks (GANs)~\cite{Goodfellow-NIPS2014} could be applied for dealing with the insufficient data problem in terms of an advanced data augmentation. In the same paper, the authors also outline another potential usage of GANs in the autonomous driving area. A more formal introduction to the soling detection and categorization is provided in~\cite{uricar2019soilingnet}. The problem is formalized as a multilabel classification task and also discusses the applications of soiling detection, including the camera cleaning. U\v{r}i\v{c}\'{a}\v{r} et al.~\cite{uricar2019desoiling} provided a desoiling dataset benchmark. 

The rest of this section provides an overview of the commonly used GAN based image-to-image translation framework, which we use in the proposed method. Due to the limited work on soiling, we also review the closely related area of adverse weather scenarios.




\subsection{GAN based Image-to-Image Translation} \label{subsec:i2it}

In recent years, the task of artificial image generation is dominated by GANs~\cite{Goodfellow-NIPS2014}, showing a great ability to synthesize realistic-looking images in~\cite{Park0WZ19}. The Image-to-Image translation is a part of graphics and computer vision that aims to learn a mapping between a~source domain $X$ and a target domain $Y$ with the use of paired data. In~\cite{IsolaZZE17}, the authors present a method using GANs to tackle the problem of image-to-image translation using paired data. However, obtaining such paired data can be difficult and sometimes even impossible. Therefore, an unsupervised version, without any examples of corresponding pairs, is even more critical and challenging.

This problem is tackled by CycleGAN~\cite{Zhu-ICCV-2017} with the use of two mappings $G\colon X \rightarrow Y$ and $F\colon Y \rightarrow X$. Since these mappings are highly under-constrained, they propose to use a \emph{cycle consistency loss} to enforce $F \left( G \left( X \right) \right) \approx X$. Even though it is not emphasized that much in the CycleGAN paper, the authors in their implementation use also \emph{identity losses}, $G(Y) \approx Y$ and $F(X) \approx X$, which improve the results significantly.

\subsection{Rainy Scenes} \label{subsec:related_water}

Rainy scenes are slightly related to water soiling situation. Because this degradation is semi-transparent and some background information is still visible, it is common to use an image restoration algorithm that improves the image's quality. The work of~\cite{Li_2019_CVPR} provides a comprehensive analysis of this topic. The authors of \cite{Yang_2019_CVPR} address the problem of rain removal from videos by a two-stage recurrent network. The rain-free image is estimated from the single rain frame at the first stage. This initial estimate serves as guidance along with previously recovered clean frames to help to obtain a~more accurate, clean frame at the second stage.

In~\cite{Ren_2019_CVPR}, the authors propose a progressive recurrent deraining network by repeatedly unfolding a shallow ResNet with a recurrent layer.
In~\cite{Wang_2019_CVPR}, a dataset of $\approx 29.5$k rain/rain-free image pairs are constructed, and a SPatial Attentive Network (SPANet) is proposed to remove rain streaks in a local-to-global manner.
Porav et al.~\cite{porav2019i} presented a method that improves the segmentation tasks on images affected by rain. They also introduced a dataset of clear-soiled image pairs, which is used to train a denoising generator that removes the effect of real water drops.
Li et al.~\cite{LiCT19} proposed a two stage algorithm incorporating depth-guided GAN for heavy rain image restoration. 
Quan et al.~\cite{QuanDCJ19} use double attention mechanism CNN for raindrop removal.

\subsection{Dehazing} \label{subsec:related_other}

Another type of image quality degradation is caused by the presence of aerosols (e.g., mist, fog, fumes, dust, \dots) in the environment surrounding the car. Due to the light scattering caused by these aerosol particles, the resulting image tends to have faint colors and looks hazy, which can inherently also impact the further image processing. 

Fattal presents in~\cite{Fattal-2008} a method for single image dehazing, based on a refined image synthesis model and a depth estimation. Berman et al.~\cite{Berman-2016-z}, on the other hand, propose a~solution, which is not based on local priors and builds on the assumption that a dehazed image can be approximated by a~few hundred distinct colors which form tight clusters in the RGB color space. Ki et al.~\cite{Ki-2018-z} propose fully end-to-end learning-based boundary equilibrium GANs to perform an ultra-high-resolution single image dehazing. Yan et al.~\cite{YanST20} propose a semi-supervised learning using a mixture of real data without ground truth and synthetic data. 




\section{Artificial Soiling Generation} \label{sec:proposed}

\begin{figure*}[tb]
    \centering 
    \fbox{\includegraphics[width=0.182\linewidth]{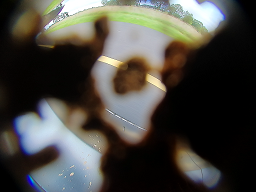}}
    \fbox{\includegraphics[width=0.182\linewidth]{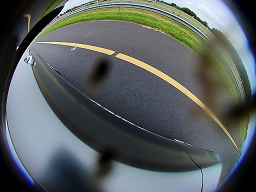}}
    \fbox{\includegraphics[width=0.182\linewidth]{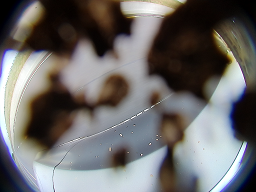}}
    \fbox{\includegraphics[width=0.182\linewidth]{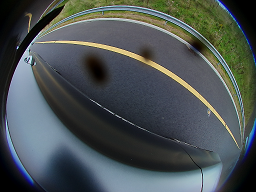}}
    \fbox{\includegraphics[width=0.182\linewidth]{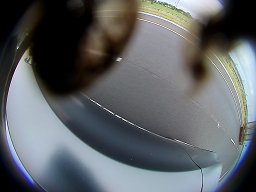}}
    
    \fbox{\includegraphics[width=0.182\linewidth]{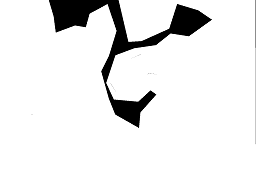}} \fbox{\includegraphics[width=0.182\linewidth]{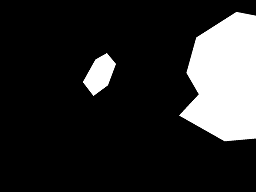}}
    \fbox{\includegraphics[width=0.182\linewidth]{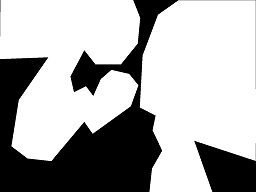}}
    \fbox{\includegraphics[width=0.182\linewidth]{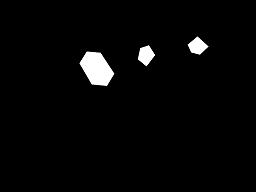}}
    \fbox{\includegraphics[width=0.182\linewidth]{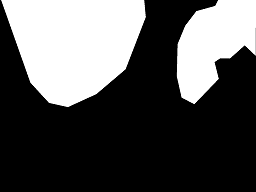}}
    \caption{Several examples from the \textit{WoodScape Dataset}. Top: RGB images from the fisheye camera. Bottom: corresponding human made annotations. White polygons represent soiling masks, while the background is marked by black color.} \label{fig:soiling_dataset}
\end{figure*}

\begin{figure*}[tb]
    \centering
    \includegraphics[width=0.4\linewidth]{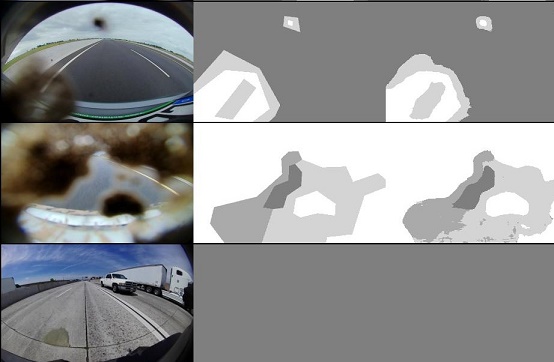}
    \includegraphics[width=0.4\linewidth]{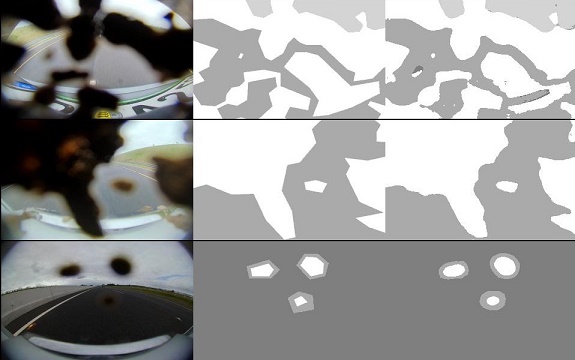}
    \caption{A few examples of the segmentation network results. In the left column are original soiled images, in the middle are the coarse annotations, and on the right column are the masks obtained by the segmentation network. The clean area is marked in dark gray, the opaque area is white, and the semi-transparent region light gray.}
    \label{fig:soiling_semantic_segmentation}
\end{figure*}

\begin{figure*}[tb]
    \centering
    \includegraphics[width=0.75\linewidth]{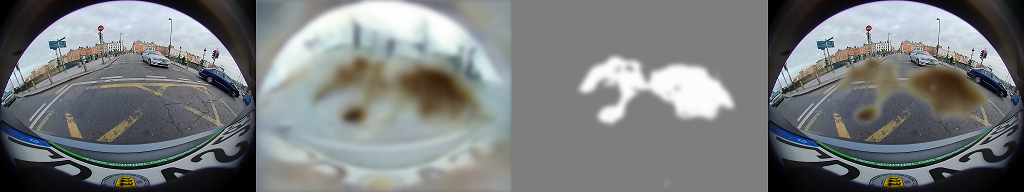}
    \caption{The soiling generation baseline pipeline. From left to right: an image into which we would like to paint a random soiling pattern; CycleGAN generated ``soiled" version; blurred mask of the segmented soiling from the generated image; the resulting artificially ``soiled" image obtained by convex combination of the original image and the generated soiling via the segmented mask.}
    \label{fig:soiling_generation_pipeline}
\end{figure*}

The task of single image soiling annotation on the fisheye cameras is quite tedious. 
We make use of polygonal annotation, which is a compromise of annotation speed and quality. However, even this kind of polygonal annotation is sometimes tough to interpret even by human annotators. It is particularly true for the soiling boundary, which is usually very fuzzy.


However, an even bigger problem is how to obtain the soiling data. In our setup, we apply a random pattern of soiling on the camera lens using a particular soiling source. Then drivers ride the car for a while and repeat the process several times. It has many limitations: Firstly, it is very inconvenient to record data for all probable scenarios (e.g., driving through the city, rural area, highway). Secondly, it is not possible to measure the real impact of soiling on the images, because we need a clean version of the same images for a fair comparison. 

All these limitations motivate us to use the synthetic soiling data. In the following sections, the proposed soiling generation algorithms are described.


\subsection{Soiling Generation Baseline Pipeline} \label{subsec:baseline_pipeline}

The core of our baseline pipeline is formed by a CycleGAN~\cite{Zhu-ICCV-2017} network, which we train to perform the image-to-image translation from clean images to their soiled counterparts. The main problem of the CycleGAN method is that it modifies the whole image. For our desired application, this can lead to undesired artifacts in the generated images. Besides that, the generated synthetic soiling patterns are often relatively realistic. Note that due to GPU memory requirements and time constraints, our CycleGAN training uses rescaled images ($\frac{1}{4}$ of both width and height). 

Next, we train a soiling semantic segmentation network, $\mathcal{M}$, using the weak polygonal annotation of soiling (see Figure~\ref{fig:soiling_dataset} for several examples). Even though the annotation is quite coarse, the segmentation network $\mathcal{M}$ is able to fit the soiling patterns more precisely. See Figure~\ref{fig:soiling_semantic_segmentation} for a~few examples of 
$\mathcal{M}$ outputs in comparison to the original annotations. 
We use \textit{WoodScape Dataset}~\cite{woodscape-2019} for training the soiling segmentation network.
Last but not least, we train a super-resolution network $\mathcal{U}$, which we use to transform the GAN generated image to the original image resolution (i.e., up-scaling of $4\times$ factor).

The idea of a baseline data generation algorithm is described in Figure~\ref{fig:soiling_generation_pipeline}. 
We take the generator transforming a~clean image to the soiled image ($\mathcal{G}_{\mathrm{C2S}}$) and apply it to the clean image $I$. It gives us an image with a random soiling pattern $I_{\mathrm{s}}$. Next, we obtain the soiling mask $m$. This is achieved by applying the semantic segmentation network on the generated soiled image followed by a Gaussian smoothing filter $\gamma$:  $m = \gamma\left(\mathcal{M}\left(I_{\mathrm{s}}\right)\right)$. The resulting soiling mask $m$ is an image with values in range $[0, 1]$, where $0$ means background, and $1$ means soiling. The intermediate values can be understood as semi-transparent soiling. We apply the Gaussian smoothing filter because it mimics the physical nature of the soiling phenomenon where the edges of the soiling patterns are typically semi-transparent, due to photon scattering.
Finally, the artificially soiled version of the original image $\hat{I}$ is a composition of the original image $I$ and the soiling pattern $I_\mathrm{s}$ via the estimated mask $m$:
\begin{equation}
    \hat{I} = \left(1 - \mathcal{U}\left(m\right) \right) \cdot I + \mathcal{U}\left(m\right) \cdot \mathcal{U}(I_{\mathrm{s}}) \;.
\end{equation}
Note, it is possible to use arbitrary images for the final composition, once we obtain $I_{\mathrm{s}}$ and $m$. The mask $m$ obtained by the semantic segmentation network $\mathcal{M}$  serves as an automatic annotation of the soiling in the generated image. 


This simple pipeline has certain limitations. The biggest one is that it cannot be expected to work smoothly for soiling types caused by water (e.g., raindrops, snow) in this specific formulation. One option for dealing with this issue is to follow the approach of~\cite{Alletto_2019_ICCV_Workshops}, where the authors model the reflection patterns of the water drops using the whole image and apply filters and transformations consequently. The other option is to formulate a CyleGAN-like approach, which can cope with changing only those parts of the image that correspond to the soiling pattern and keep the rest unchanged. We formulate this approach in the following section. 


\subsection{DirtyGAN} \label{subsec:dirtygan}




\begin{figure*}[tb]
    \centering
    \includegraphics[width=0.8\linewidth]{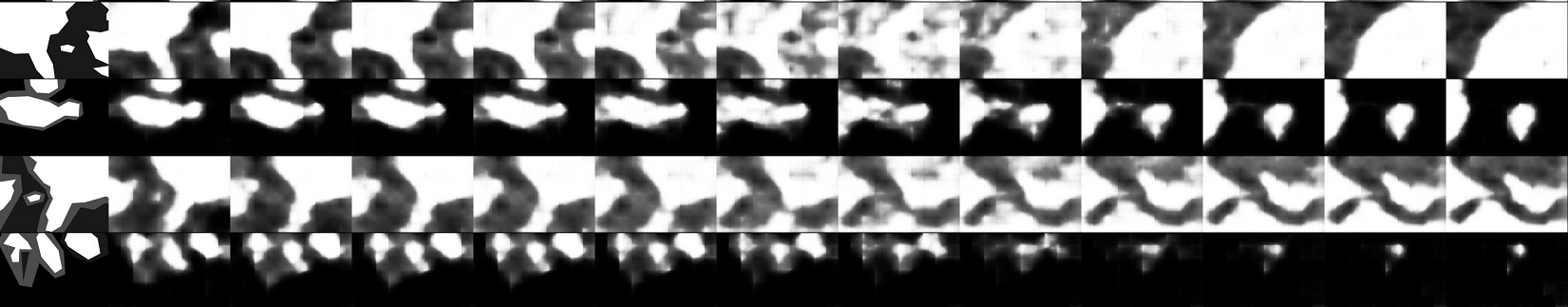}
    \caption{Variational AutoEncoder and the walk on the soiling manifold. The leftmost column depicts the original soiling pattern (the annotation for some particular soiled image). The next column is the reconstructed version by applying the whole VAE. The next $10$ columns represent the transition (the walk on the soiling manifold) from the leftmost image to the rightmost one, which represents another example.}
    \label{fig:VAE}
\end{figure*}

The problem of applying CycleGAN in artificial soiling synthesis is twofold. Firstly, CycleGAN does not constrain the image generation to any specific regions and instead regenerates the whole image, affecting all pixels. In the case of the artificial soiling generation, this is highly undesirable. For the investigation of the soiling impact on the further image processing, the background (i.e., regions of the image not affected by the soiling) must remain untouched. Secondly, the generation branch ``clean''~$\rightarrow$~``soiled'' is ill-posed, as there is no visual clue for where the soiling should be produced. There are infinitely many patterns that could be created. Furthermore, there is also no control over the soiling pattern production process.

The first problem can be addressed, e.g., by InstaGAN~\cite{MoCS19}. However, in such a case, the second problem becomes even a more significant issue. We decided to guide the pattern generation process via a Variational AutoEncoder (VAE)~\cite{Doersch16} and modify the CycleGAN algorithm so that it applies only on the masked regions of the source and target domain images. We coin the proposed network DirtyGAN. 

We use the weak polygonal soiling annotations from the \textit{WoodScape Dataset} for training the VAE. The main idea of using VAE for the soiling patterns generation is as follows. By using the encoder of the trained VAE, we can obtain the projection of an actual sample from the dataset to a lower-dimensional representation. If we select two samples $\#z_1$ and $\#z_2$ that are close on the soiling pattern manifold, we can obtain a novel sample $\#z$ by taking their convex combination.
\begin{equation}
    \#z = \alpha \#z_1 + (1-\alpha) \#z_2 \;,
\end{equation}
where $\alpha \in [0, 1]$. 
Then, we take this intermediate representation $\#z$ and apply the trained decoder from VAE to reconstruct the corresponding soiling pattern. In Figure~\ref{fig:VAE}, we depict several examples of this intermediate soiling pattern reconstruction.

\begin{figure*}[tb]
    \centering 
    \includegraphics[height=4cm, width=\linewidth]{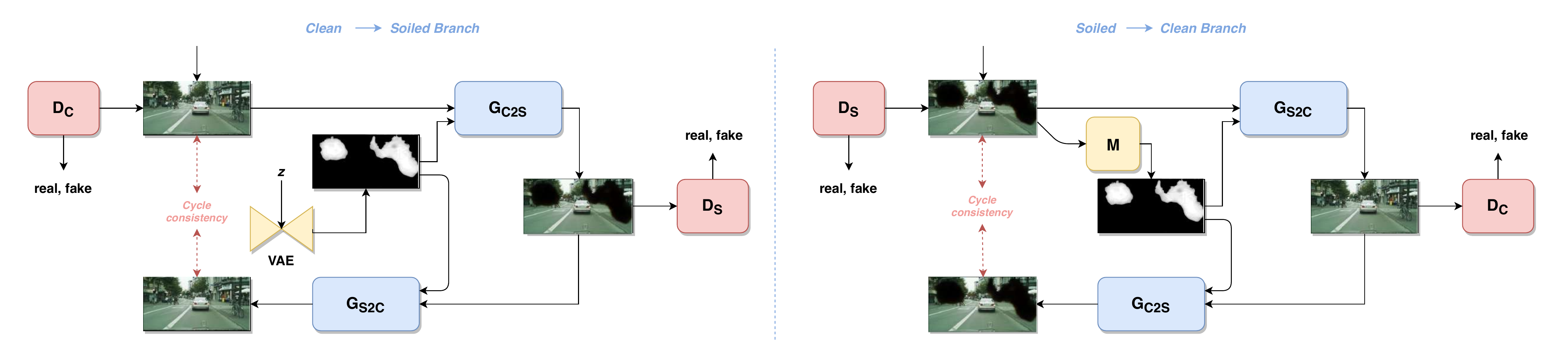}
    \caption{The DirtyGAN scheme. The original CycleGAN scheme is enhanced by VAE for novel soiling pattern generation in ``clean'' $\rightarrow$ ``soiled'' branch and by mask estimation in the opposite branch. Please see the supplementary material for a higher resolution version of this image.} \label{fig:DirtyGAN_scheme}
\end{figure*}

The benefit of using sampling from the learned VAE is that we could even use it to create animated masks, e.g., to mimic dynamic soiling effects, such as water drops in heavy rain or to be able to investigate the impact of dynamic soiling in general. 
After training the VAE, we limit CycleGAN to be applied only on the masked regions corresponding to the generated mask for the ``clean'' $\rightarrow$ ``soiled'' translation or the mask obtained by the soiling semantic segmentation mask $\mathcal{M}$. We use a similar composition as in the baseline presented in Section~\ref{subsec:baseline_pipeline}. In Figure~\ref{fig:DirtyGAN_scheme}, we depict the whole DirtyGAN scheme.

\subsection{Dirty Datasets} \label{subsec:dirty_woodscape}

We have used the baseline pipeline to generate artificial soiling on our recently published \textit{WoodScape Dataset}~\cite{woodscape-2019} with $10$k images, which comes with semantic segmentation annotation. It makes it a suitable candidate for the soiling generation since we can merge the provided annotation with the soiling mask and measure the direct impact on classification models. Our generated data with updated annotation will be released as a \textit{WoodScape Dataset} companion under the name of \textit{Dirty WoodScape Dataset}\footnote{\url{https://github.com/valeoai/WoodScape}}.
 
\begin{figure*}[tb]
    \centering 
    \fbox{\includegraphics[width=0.182\linewidth]{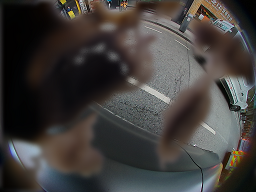}}
    \fbox{\includegraphics[width=0.182\linewidth]{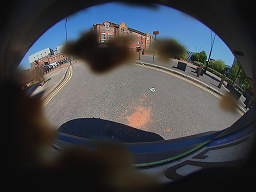}}
    \fbox{\includegraphics[width=0.182\linewidth]{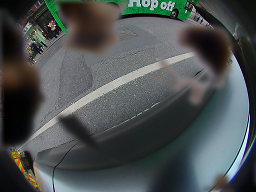}}
    \fbox{\includegraphics[width=0.182\linewidth]{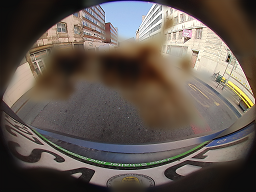}}
    \fbox{\includegraphics[width=0.182\linewidth]{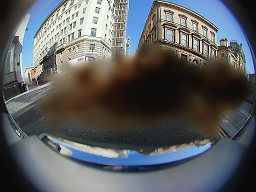}}
    
    \fbox{\includegraphics[width=0.182\linewidth]{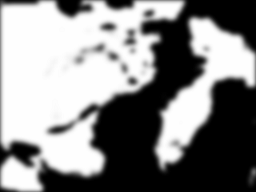}} \fbox{\includegraphics[width=0.182\linewidth]{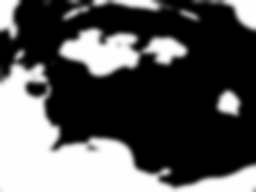}}
    \fbox{\includegraphics[width=0.182\linewidth]{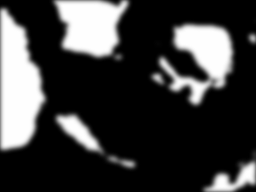}}
    \fbox{\includegraphics[width=0.182\linewidth]{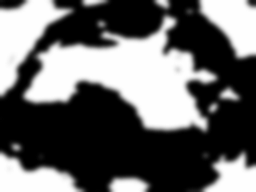}}
    \fbox{\includegraphics[width=0.182\linewidth]{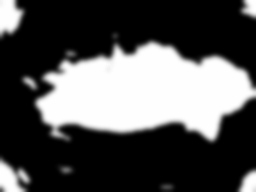}}
    \caption{Examples of the generated images from the \textit{DirtyWoodScape Dataset}, together with the generated annotations. Top: \textit{WoodScape Dataset} RGB images with generated opaque soiling. Bottom: corresponding automatically generated annotations. White pixels represent the soiling, while black pixels represent the background.} \label{fig:dirty_dataset_examples}
    \vspace{-4mm}
\end{figure*}

In Figure~\ref{fig:dirty_dataset_examples}, we show several generated examples together with their automatically obtained annotations. Our method of soiling generation is not limited to fisheye images only. Since we would like to support standard benchmarking as well, we also release a Dirty Cityscapes dataset. It is, as the name suggests, based on the Cityscapes \cite{cordts2016cityscapes} dataset.

\section{Experimental Evaluation} \label{sec:experiments}

\subsection{Data Augmentation for Soiling Detection Task}

A primary purpose of generating soiling is to mimic the real soiling scenarios. 
Firstly, we trained a soiling segmentation network on real soiling data only ($8$k images). We tested this model performance on real soiling test data 
and achieved $73.95\%$ accuracy. Then, we added 
generated soiling data ($8$k images) to our training catalog. As the generated soiling replicates the real soiling patterns, the network's performance increased to $91.71\%$ on real soiling test data, forming a $17.76\%$ increase in accuracy without the need for costly annotations and real-time soiling scene captures  
(see the second and fourth row of Table~\ref{tab:quantitative_comparison}). 

We observe a significant reduction in accuracy when the network is trained on artificial images solely. It can be attributed to a limited ability to capture the entire data diversity to support the original soiled data. We also conduct a simple ablation study with standard data augmentation techniques (flipping, contrast changes) to match the size of training data with generated samples and observe a much lesser improvement in accuracy (third row), compared to the scenario with augmentation by generated synthetic samples.

The classifier used in the experiment uses 
ResNet50~\cite{he2016deep} for an encoder and FCN8~\cite{Shelhamer-2017} for a decoder. The binary cross-entropy was used as a loss function with the ADAM optimizer~\cite{kingma2014adam} for training with a learning rate of $1\times10^{-4}$. The image resolution was $640 \times 480$ pixels. 

\begin{table}[tbh]
    \centering
    \caption{Comparison of Soiling Segmentation model trained on generated and real soiled images. Accuracy is computed on a real test dataset with $2,000$ images.}
    \vspace{0.25cm}
    \begin{adjustbox}{width=0.46\textwidth}
    \begin{tabular}{l|r}
    \multicolumn{1}{c|}{\textbf{\begin{tabular}[c]{@{}c@{}}ResNet50-FCN8 based\\ Soiling Segmentation Model\end{tabular}}} & \multicolumn{1}{c}{\textbf{\begin{tabular}[c]{@{}c@{}}Accuracy [\%] on real \\ test 2K dataset (mIoU)\end{tabular}}} \\ 
    \hline
    \begin{tabular}[c]{@{}l@{}}Trained solely on generated images\\ (8000 samples)\end{tabular} & 47.41 \\ 
    \begin{tabular}[c]{@{}l@{}}Trained solely on real images\\ (8000 samples)\end{tabular} & 73.95 \\ 
    \begin{tabular}[c]{@{}l@{}}Trained on real \& data augmentation\\ (16000 samples)\end{tabular} & 78.20 \\ 
    \begin{tabular}[c]{@{}l@{}}Trained on real \& generated images\\ (16000 samples)\end{tabular} & \textbf{91.71} \\ 
    \end{tabular}
    \label{tab:quantitative_comparison}
    \end{adjustbox}
\end{table}

\subsection{Artificial Soiling Quality} \label{subsec:soiling_quality}

We performed a subjective visual study of the quality of the artificial soiling similarly as it is done in other GAN based image generation algorithms. We selected representative human participants with a variable level of the problem understanding: ranging from absolutely no knowledge about the computer vision and surround-view camera imaging systems for autonomous driving to people working with the soiling data daily. The participants were asked to classify the presented images either as real ones or fakes. Then, we randomly showed images with the real soiling from the \textit{WoodScape Dataset}~\cite{woodscape-2019} and with the artificially generated soiling. To make the task even more complicated, we used soiled images showing similar scenes as they occur in the \textit{WoodScape Dataset}. Otherwise, the participants might eventually spot the differences in the data distribution as the real soiling data comes only from limited scenarios.

\begin{table}[ht!]
    \centering
    \caption{Quantitative evaluation on our \textit{WoodScape} dataset using DeepLabV3+. The accuracy measure is mIoU [\%].}
    \begin{tabular}{l|r|r}
    \textbf{Train\textbackslash Test} & \textbf{Clean} & \textbf{Soiled} \\ \hline
    Clean & $56.6$	& $34.8$ \\ 
    Soiled & $52.1$	& $48.2$	\\ 
    \end{tabular}
    \label{table:DeepLabV3}
\end{table}

\vspace{-4mm}

\begin{table}[ht!]
    \centering
    \caption{Quantitative evaluation on Cityscapes using DeepLabV3+. The accuracy measure is mIoU [\%].}
    \begin{tabular}{l|r|r}
    \textbf{Train\textbackslash Test} & \textbf{Clean} & \textbf{Soiled} \\ \hline
    Clean & $38.1$	& $26.6$ \\ 
    Soiled & $35.5$	& $38.0$ \\ 
    \end{tabular}
    \label{table:DeepLabV3Cityscapes}
\end{table}

\begin{figure*}[t!]
\centering
\begin{subfigure}{.2\textwidth}
    \includegraphics[width=\textwidth]{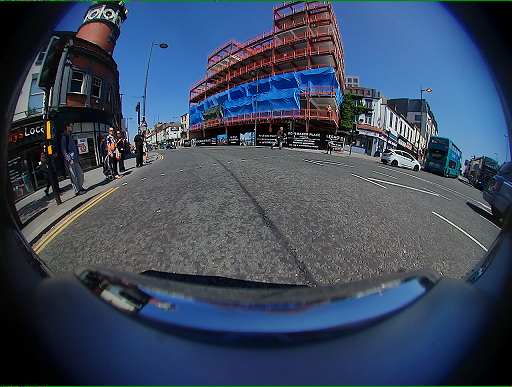}
    \caption{\textcolor{black}{Clean Image}\vspace{0.3mm}}
    \label{fig:deeplab_a}
\end{subfigure}%
\quad
\begin{subfigure}{.2\textwidth}
    \includegraphics[width=\textwidth]{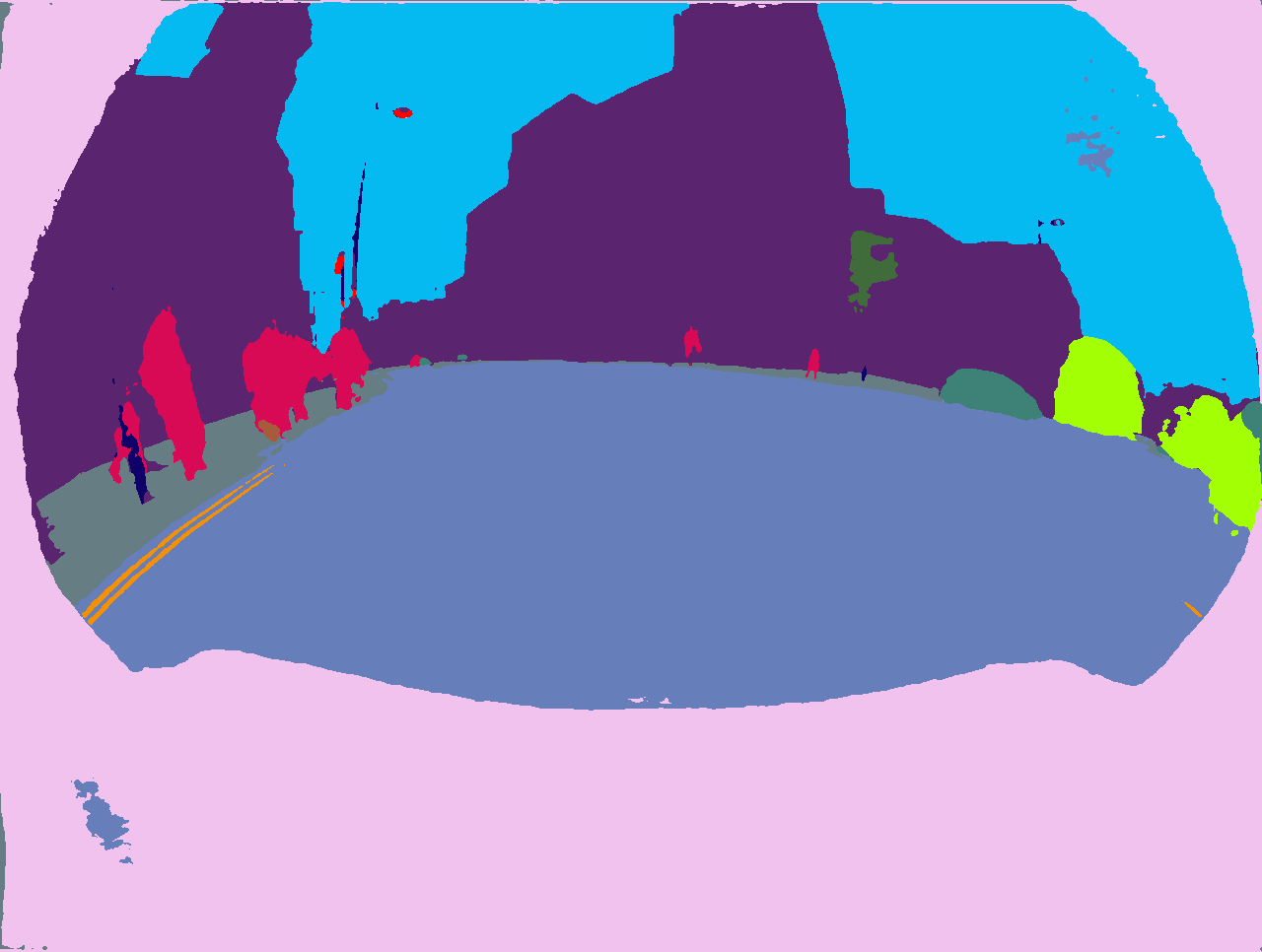}
    \caption{\textcolor{black}{Train clean, test clean}\vspace{0.3mm}}
    \label{fig:deeplab_b}
\end{subfigure}%
\quad
\begin{subfigure}{.2\textwidth}
    \includegraphics[width=\textwidth]{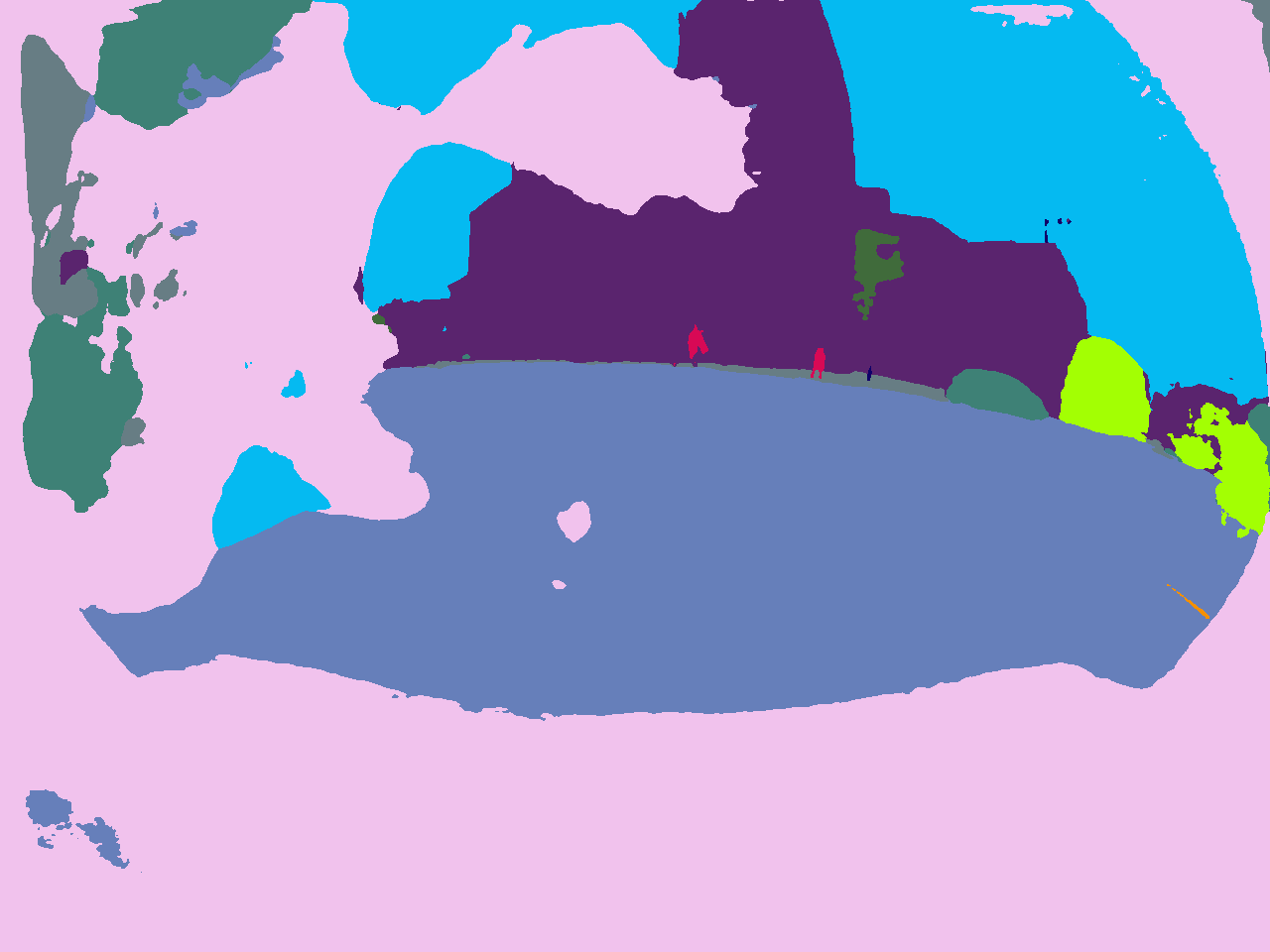}
    \caption{\textcolor{black}{Train clean, test soil}\vspace{0.3mm}}
    \label{fig:deeplab_c}
\end{subfigure}%
\quad
\begin{subfigure}{.2\textwidth}
    \includegraphics[width=\textwidth]{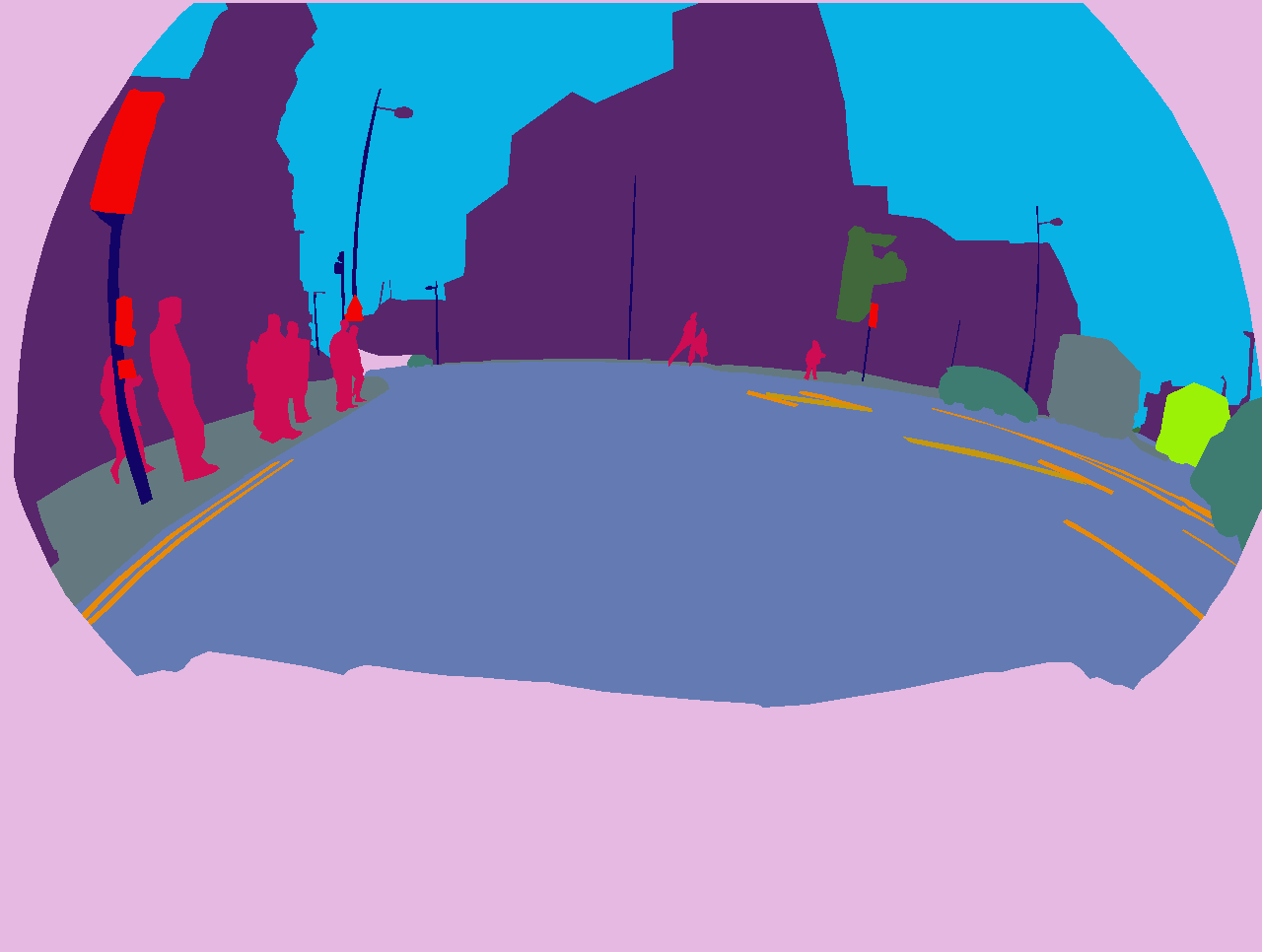}
    \caption{\textcolor{black}{Ground Truth}\vspace{0.3mm}}
    \label{fig:deeplab_d}
\end{subfigure}%
\newline
\begin{subfigure}{.2\textwidth}
    \includegraphics[width=\textwidth]{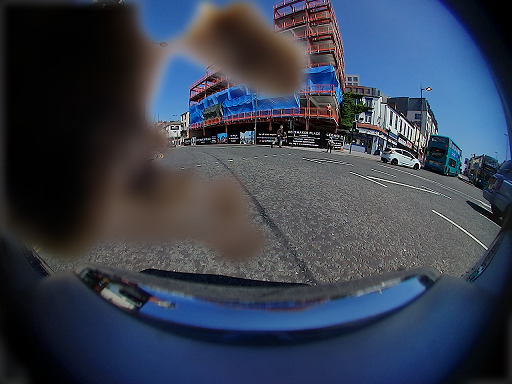}
    \caption{\textcolor{black}{Soiled Image}}
    \label{fig:deeplab_e}
\end{subfigure}%
\quad
\begin{subfigure}{.2\textwidth}
    \includegraphics[width=\textwidth]{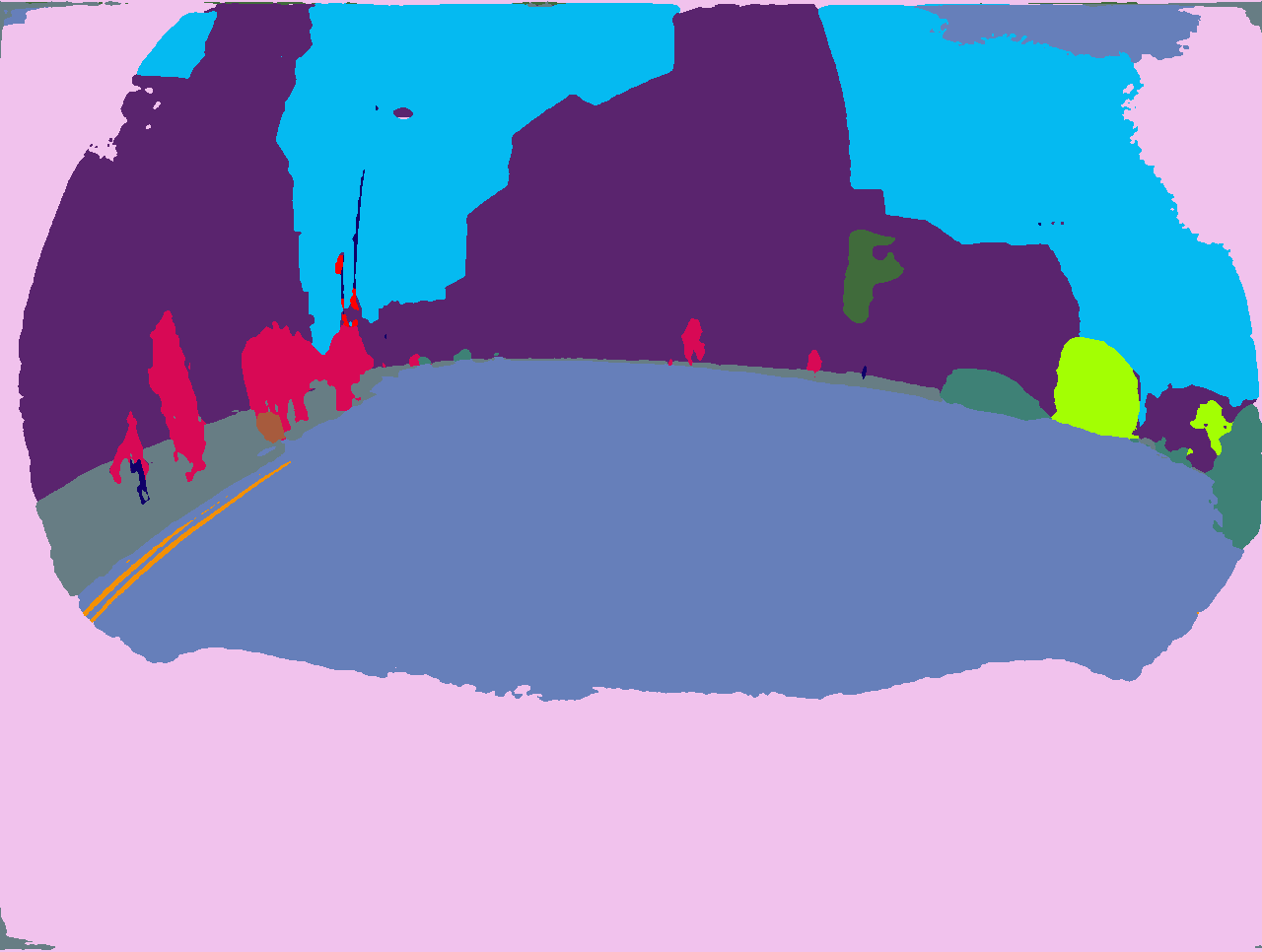}
    \caption{\textcolor{black}{Train soil, test clean}}
    \label{fig:deeplab_f}
\end{subfigure}%
\quad
\begin{subfigure}{.2\textwidth}
    \includegraphics[width=\textwidth]{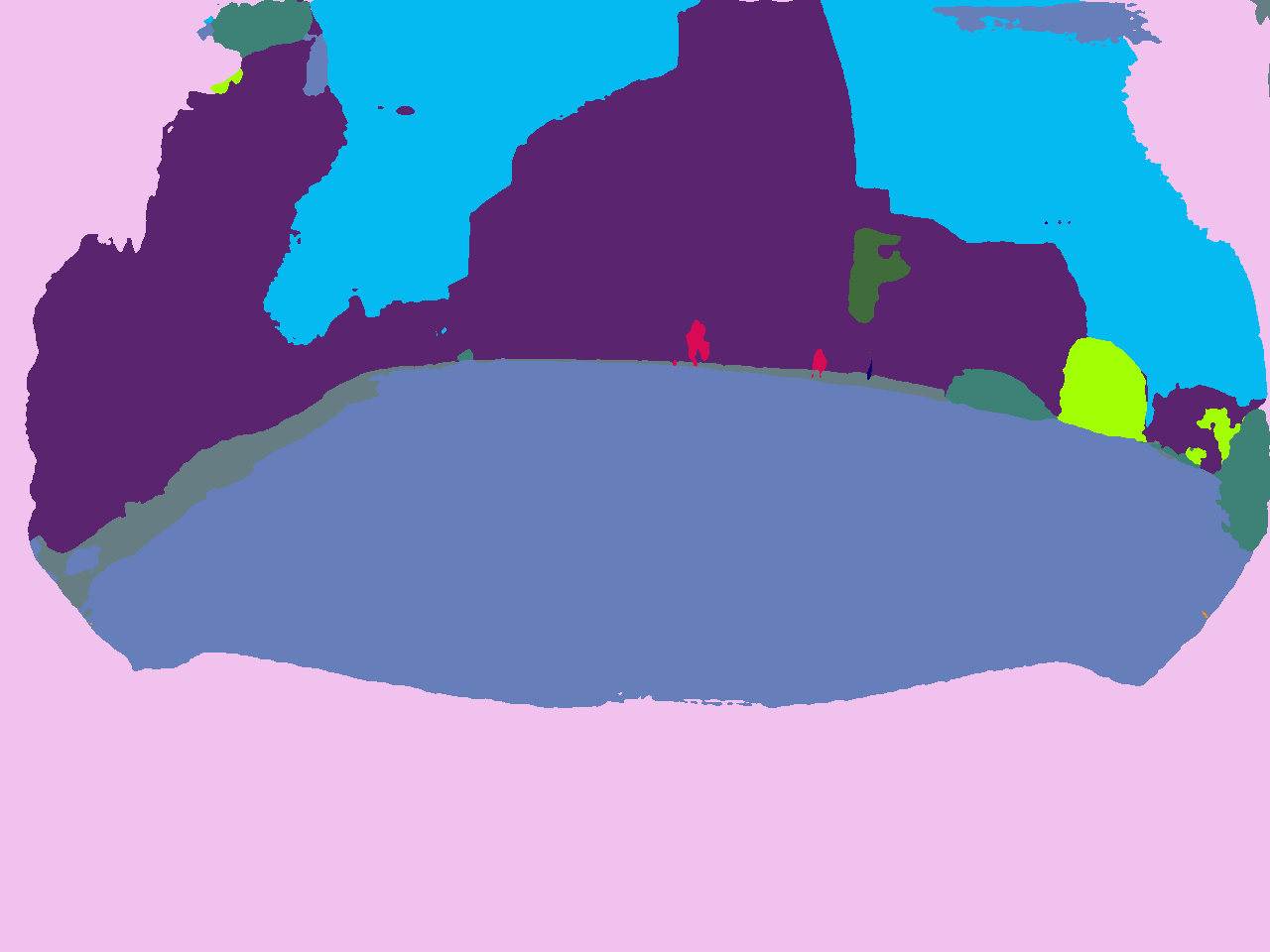}
    \caption{\textcolor{black}{Train soil, test soil}}
    \label{fig:deeplab_g}
\end{subfigure}%
\vspace{-2mm}
\caption{Qualitative evaluation using clean vs soiled images on the semantic segmentation network DeepLabV3+.}
\label{fig:deeplab_1}
\end{figure*}

\begin{figure*}[htpb]
    \centering 
    \includegraphics[height=3cm, width=0.9\linewidth]{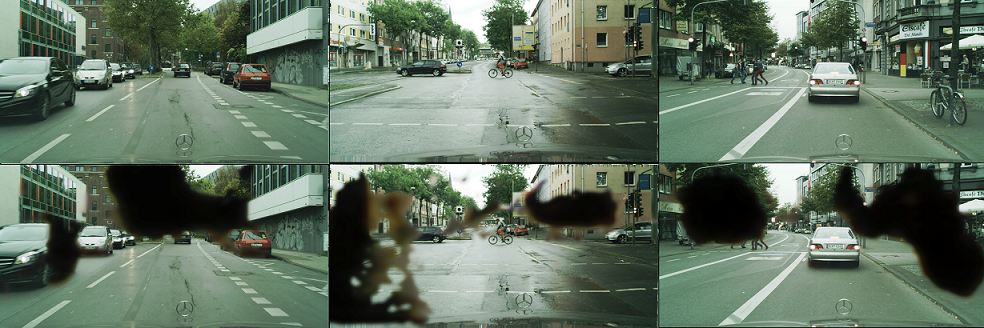}
    \caption{Examples of the generated images from the \textit{Dirty Cityscapes Dataset}. Top: \textit{Cityscapes Dataset} RGB images. Bottom: Generated soiled images corresponding to clean images.} \label{fig:DirtyCityscapes}
    \vspace{-4mm}
\end{figure*}

The non-expert participants were not able to recognize real images from fakes. The expert participants were sometimes able to spot a difference in the soiling pattern because it was novel, and also they sometimes spotted small artifacts, e.g., blurry texture. However, expert participants were biased as they know the real data. In general, we can say that the image quality of the generated artificial soiling is satisfactory when judged by human inspection \textcolor{black}{for $95\%$ of scenarios.}

We chose a random subset of $5\%$ of generated samples for quantitative comparison due to the manual annotation cost. We use the mIoU metric to compare between artificially generated annotation and manual annotation. We obtain a high accuracy of $93.2\%$. It is to be noted that manual annotation can be subjected, especially at boundaries, and sometimes it is worse than artificial annotation based on visual inspection.


\subsection{Degradation Effect on Semantic Segmentation} \label{subsec:degradation}

To demonstrate the impact of the soiling on the camera lens, we provide an empirical evaluation on semantic segmentation task. Pixel level classification is commonly used in autonomous driving applications. Our dataset consists of $10$k pairs of images (clean and synthetically soiled using the proposed baseline framework).
We split our data into training and testing by $80:20$ ratio according to the sampling strategy described in~\cite{Uricar-2019b}. 
We trained two DeepLabV3+~\cite{chen2018encoder} models on the clean and soiled images, respectively. We evaluate the performance separately on clean and soiled test data. Table~\ref{table:DeepLabV3} summarizes the obtained results. 

A segmentation model trained on clean images records $56.6\%$ mIoU on clean test data and $34.8\%$ on soiled data, a performance drop of $21.8\%$ compared to clean images. 
This significant drop shows that soiling can cause severe degradation to a standard visual perception task in autonomous driving. 

A model trained on the synthetic soiling data shows a limited degradation to $16.1\%$.
However, training on the soiled images shows a $4\%$ accuracy degradation on clean test data compared to the baseline when evaluated on clean images. 
Note, that model trained on the synthetic data perform reasonably well on the clean data with a few percentage drop in mIoU. This is expected as the data contain additional class, which is treated as background/unknown (void class). Thus, the trained model used less portion of data to train the remaining classes.

Figure~\ref{fig:deeplab_1} 
depicts a qualitative evaluation 
of the soiling impact on the segmentation task.
The baseline model trained on clean images \ref{fig:deeplab_a} is evaluated on soiled images in \ref{fig:deeplab_c} showing a high level of degradation due to the soiling. Figure~\ref{fig:deeplab_d} shows the ground truth annotations, while \ref{fig:deeplab_f} and \ref{fig:deeplab_g} illustrate results of model trained on the soiled images while testing on the clean images in \ref{fig:deeplab_a} and soiled images in \ref{fig:deeplab_e}. 
In a realistic scenario, annotations are not available for the occluded region of soiled images. Using our GAN generated dataset, we use annotations in the soiled area to enable the model to interpolate segmentation classes in occluded soiled parts. Figure~\ref{fig:deeplab_f} shows the capability of segmentation networks to perform segmentation even behind the soiled area. However, it is less reliable compared to the clean baseline and sensitive to overfitting.

The same type of experiments were conducted using the Cityscapes dataset~\cite{cordts2016cityscapes}. The results are presented in Table~\ref{table:DeepLabV3Cityscapes}, which, as you can see, show a similar trend as in the WoodScape experiment.


\section{Conclusions} \label{sec:conclusions}


In this paper, we proposed two algorithms for the generation of soiling on images from surround view fisheye cameras. The first algorithm is a pipeline built from several well-known blocks, such as CycleGAN~\cite{Zhu-ICCV-2017}, semantic segmentation of the generated soiling, and image composition. The second algorithm is a novel DirtyGAN network, which can generate similar results as the baseline pipeline in an end-to-end fashion. 
The possibility to generate random but realistic soiling patterns on camera images is an integral component in examining the degradation of other image processing methods. We provided an empirical evaluation of the performance degradation on several typical classification tasks common for autonomous driving scenarios. We demonstrate that our soiling model trained on fisheye images generalizes well on the Cityscapes dataset, enabling us to create dirty versions of public datasets. Last but not least, we release a public dataset as a companion to the recently published \textit{WoodScape Dataset}~\cite{woodscape-2019}, coined \textit{Dirty WoodScape Dataset}, which can serve as a benchmark for measuring the degradation of the off-the-shelf classification algorithms.


\paragraph{Acknowledgements.}
Anton\'{i}n Vobeck\'{y} was supported by the Grant Agency of the Czech Technical University in Prague, grant No. SGS18/205/OHK3/3T/37. All authors would like to thank Valeo for the support of publishing.

{\small
\bibliographystyle{ieee_fullname}
\bibliography{bib/references}

\begin{thebibliography}{10}\itemsep=-1pt

\bibitem{Alletto_2019_ICCV_Workshops}
Stefano Alletto, Casey Carlin, Luca Rigazio, Yasunori Ishii, and Sotaro
  Tsukizawa.
\newblock Adherent raindrop removal with self-supervised attention maps and
  spatio-temporal generative adversarial networks.
\newblock In {\em The IEEE International Conference on Computer Vision (ICCV)
  Workshops}, 2019.

\bibitem{Berman-2016-z}
Dana Berman, Tali Treibitz, and Shai Avidan.
\newblock Non-local image dehazing.
\newblock In {\em 2016 {IEEE} Conference on Computer Vision and Pattern
  Recognition, {CVPR} 2016, Las Vegas, NV, USA, June 27-30, 2016}, 2016.

\bibitem{chen2018encoder}
Liang-Chieh Chen, Yukun Zhu, George Papandreou, Florian Schroff, and Hartwig
  Adam.
\newblock Encoder-decoder with atrous separable convolution for semantic image
  segmentation.
\newblock In {\em Proceedings of the European conference on computer vision
  (ECCV)}, 2018.

\bibitem{cordts2016cityscapes}
Marius Cordts, Mohamed Omran, Sebastian Ramos, Timo Rehfeld, Markus Enzweiler,
  Rodrigo Benenson, Uwe Franke, Stefan Roth, and Bernt Schiele.
\newblock The cityscapes dataset for semantic urban scene understanding.
\newblock In {\em Proceedings of the IEEE conference on computer vision and
  pattern recognition}, 2016.

\bibitem{Doersch16}
Carl Doersch.
\newblock Tutorial on variational autoencoders.
\newblock In {\em CoRR}, 2016.

\bibitem{Fattal-2008}
Raanan Fattal.
\newblock Single image dehazing.
\newblock In {\em {ACM} Trans. Graph.}, 2008.

\bibitem{Goodfellow-NIPS2014}
Ian Goodfellow, Jean Pouget-Abadie, Mehdi Mirza, Bing Xu, David Warde-Farley,
  Sherjil Ozair, Aaron Courville, and Yoshua Bengio.
\newblock Generative adversarial nets.
\newblock In {\em Advances in neural information processing systems}, 2014.

\bibitem{he2016deep}
Kaiming He, Xiangyu Zhang, Shaoqing Ren, and Jian Sun.
\newblock {Deep Residual Learning for Image Recognition}.
\newblock In {\em Proc. of CVPR}, 2016.

\bibitem{heimberger2017computer}
Markus Heimberger, Jonathan Horgan, Ciar{\'a}n Hughes, John McDonald, and
  Senthil Yogamani.
\newblock Computer vision in automated parking systems: Design, implementation
  and challenges.
\newblock In {\em Image and Vision Computing}. Elsevier, 2017.

\bibitem{SAE_automation}
S.~A.~E. International.
\newblock Taxonomy and definitions for terms related to on-road motor vehicle
  automated driving systems {J3016}, 2018.

\bibitem{IsolaZZE17}
Phillip Isola, Jun{-}Yan Zhu, Tinghui Zhou, and Alexei~A. Efros.
\newblock Image-to-image translation with conditional adversarial networks.
\newblock In {\em 2017 {IEEE} Conference on Computer Vision and Pattern
  Recognition, {CVPR} 2017, Honolulu, HI, USA, July 21-26, 2017}, 2017.

\bibitem{Ki-2018-z}
Sehwan Ki, Hyeonjun Sim, Jae{-}Seok Choi, Saehun Kim, and Munchurl Kim.
\newblock Fully end-to-end learning based conditional boundary equilibrium
  {GAN} with receptive field sizes enlarged for single ultra-high resolution
  image dehazing.
\newblock In {\em 2018 {IEEE} Conference on Computer Vision and Pattern
  Recognition Workshops, {CVPR} Workshops 2018, Salt Lake City, UT, USA, June
  18-22, 2018}, 2018.

\bibitem{kingma2014adam}
Diederik~P. Kingma and Jimmy Ba.
\newblock {Adam: A Method for Stochastic Optimization}.
\newblock In {\em Proc. of ICLR}, 2015.

\bibitem{LiCT19}
Ruoteng Li, Loong{-}Fah Cheong, and Robby~T. Tan.
\newblock Heavy rain image restoration: Integrating physics model and
  conditional adversarial learning.
\newblock In {\em {IEEE} Conference on Computer Vision and Pattern Recognition,
  {CVPR} 2019, Long Beach, CA, USA, June 16-20, 2019}, pages 1633--1642.
  Computer Vision Foundation / {IEEE}, 2019.

\bibitem{Li_2019_CVPR}
Siyuan Li, Iago~Breno Araujo, Wenqi Ren, Zhangyang Wang, Eric~K. Tokuda,
  Roberto~Hirata Junior, Roberto Cesar-Junior, Jiawan Zhang, Xiaojie Guo, and
  Xiaochun Cao.
\newblock Single image deraining: A comprehensive benchmark analysis.
\newblock In {\em The IEEE Conference on Computer Vision and Pattern
  Recognition (CVPR)}, 2019.

\bibitem{maddu2019fisheyemultinet}
Pullaro Maddu, Wayne Doherty, Ganesh Sistu, Isabelle Leang, Michal Uricar,
  Sumanth Chennupati, Hazem Rashed, Jonathan Horgan, Ciaran Hughes, and Senthil
  Yogamani.
\newblock Fisheyemultinet: Real-time multi-task learning architecture for
  surround-view automated parking system.
\newblock In {\em Proceedings of the Irish Machine Vision and Image Processing
  Conference}, 2019.

\bibitem{MoCS19}
Sangwoo Mo, Minsu Cho, and Jinwoo Shin.
\newblock Instagan: Instance-aware image-to-image translation.
\newblock In {\em 7th International Conference on Learning Representations,
  {ICLR} 2019, New Orleans, LA, USA, May 6-9, 2019}, 2019.

\bibitem{Park0WZ19}
Taesung Park, Ming{-}Yu Liu, Ting{-}Chun Wang, and Jun{-}Yan Zhu.
\newblock Semantic image synthesis with spatially-adaptive normalization.
\newblock In {\em {IEEE} Conference on Computer Vision and Pattern Recognition,
  {CVPR} 2019, Long Beach, CA, USA, June 16-20, 2019}, 2019.

\bibitem{porav2019i}
Horia Porav, Tom Bruls, and Paul Newman.
\newblock I can see clearly now : Image restoration via de-raining.
\newblock In {\em CoRR}, 2019.

\bibitem{QuanDCJ19}
Yuhui Quan, Shijie Deng, Yixin Chen, and Hui Ji.
\newblock Deep learning for seeing through window with raindrops.
\newblock In {\em 2019 {IEEE/CVF} International Conference on Computer Vision,
  {ICCV} 2019, Seoul, Korea (South), October 27 - November 2, 2019}, pages
  2463--2471. {IEEE}, 2019.

\bibitem{Ren_2019_CVPR}
Dongwei Ren, Wangmeng Zuo, Qinghua Hu, Pengfei Zhu, and Deyu Meng.
\newblock Progressive image deraining networks: A better and simpler baseline.
\newblock In {\em The IEEE Conference on Computer Vision and Pattern
  Recognition (CVPR)}, 2019.

\bibitem{Shelhamer-2017}
Evan Shelhamer, Jonathan Long, and Trevor Darrell.
\newblock Fully convolutional networks for semantic segmentation.
\newblock In {\em {IEEE} Trans. Pattern Anal. Mach. Intell.}, 2017.

\bibitem{siam2017deep}
Mennatullah Siam, Sara Elkerdawy, Martin Jagersand, and Senthil Yogamani.
\newblock Deep semantic segmentation for automated driving: Taxonomy, roadmap
  and challenges.
\newblock In {\em 2017 IEEE 20th International Conference on Intelligent
  Transportation Systems (ITSC)}, 2017.

\bibitem{siam2018rtseg}
Mennatullah Siam, Mostafa Gamal, Moemen Abdel-Razek, Senthil Yogamani, and
  Martin Jagersand.
\newblock Rtseg: Real-time semantic segmentation comparative study.
\newblock In {\em 2018 25th IEEE International Conference on Image Processing
  (ICIP)}, 2018.

\bibitem{siam2018modnet}
Mennatullah Siam, Heba Mahgoub, Mohamed Zahran, Senthil Yogamani, Martin
  Jagersand, and Ahmad El-Sallab.
\newblock {MODNet}: Motion and appearance based moving object detection network
  for autonomous driving.
\newblock In {\em Proceedings of the 21st International Conference on
  Intelligent Transportation Systems (ITSC)}, 2018.

\bibitem{Uricar-2019b}
Michal U{\v{r}}i{\v{c}}{\'{a}}{\v{r}}, David Hurych, Pavel
  K{\v{r}}{\'{\i}}{\v{z}}ek, and Senthil Yogamani.
\newblock Challenges in designing datasets and validation for autonomous
  driving.
\newblock In {\em Proceedings of the 14th International Joint Conference on
  Computer Vision, Imaging and Computer Graphics Theory and Applications -
  Volume 5: VISAPP,}. SciTePress, 2019.

\bibitem{Uricar-2019a}
Michal U{\v{r}}i{\v{c}}{\'a}{\v{r}}, Pavel K{\v{r}}{\'{i}}{\v{z}}ek, David
  Hurych, Ibrahim Sobh, Senthil Yogamani, and Patrick Denny.
\newblock {Yes, we GAN:} applying adversarial techniques for autonomous
  driving.
\newblock In {\em Electronic Imaging}, 2019.

\bibitem{uricar2019desoiling}
Michal Uric{\'a}r, Jan Ulicny, Ganesh Sistu, Hazem Rashed, Pavel Krizek, David
  Hurych, Antonin Vobecky, and Senthil Yogamani.
\newblock Desoiling dataset: Restoring soiled areas on automotive fisheye
  cameras.
\newblock In {\em Proceedings of the IEEE International Conference on Computer
  Vision Workshops}, 2019.

\bibitem{uricar2019soilingnet}
Michal U\v{r}i\v{c}\'{a}\v{r}, Pavel K\v{r}\'{i}\v{z}ek, Ganesh Sistu, and
  Senthil Yogamani.
\newblock Soilingnet: Soiling detection on automotive surround-view cameras.
\newblock In {\em 2019 22nd International Conference on Intelligent
  Transportation Systems (ITSC)}, 2019.

\bibitem{Wang_2019_CVPR}
Tianyu Wang, Xin Yang, Ke Xu, Shaozhe Chen, Qiang Zhang, and Rynson~W.H. Lau.
\newblock Spatial attentive single-image deraining with a high quality real
  rain dataset.
\newblock In {\em The IEEE Conference on Computer Vision and Pattern
  Recognition (CVPR)}, 2019.

\bibitem{YanST20}
Wending Yan, Aashish Sharma, and Robby~T. Tan.
\newblock Optical flow in dense foggy scenes using semi-supervised learning.
\newblock In {\em 2020 {IEEE/CVF} Conference on Computer Vision and Pattern
  Recognition, {CVPR} 2020, Seattle, WA, USA, June 13-19, 2020}, pages
  13256--13265. {IEEE}, 2020.

\bibitem{Yang_2019_CVPR}
Wenhan Yang, Jiaying Liu, and Jiashi Feng.
\newblock Frame-consistent recurrent video deraining with dual-level flow.
\newblock In {\em The IEEE Conference on Computer Vision and Pattern
  Recognition (CVPR)}, 2019.

\bibitem{woodscape-2019}
Senthil Yogamani, Ciar{\'{a}}n Hughes, Jonathan Horgan, Ganesh Sistu, Padraig
  Varley, Derek O'Dea, Michal U\v{r}i\v{c}{\'{a}}\v{r}, Stefan Milz, Martin
  Simon, Karl Amende, Christian Witt, Hazem Rashed, Sumanth Chennupati, Sanjaya
  Nayak, Saquib Mansoor, Xavier Perroton, and Patrick Perez.
\newblock {WoodScape: {A} multi-task, multi-camera fisheye dataset for
  autonomous driving}.
\newblock In {\em The IEEE International Conference on Computer Vision (ICCV)},
  2019.

\bibitem{Zhu-ICCV-2017}
Jun{-}Yan Zhu, Taesung Park, Phillip Isola, and Alexei~A. Efros.
\newblock {Unpaired Image-to-Image Translation Using Cycle-Consistent
  Adversarial Networks}.
\newblock In {\em {IEEE} International Conference on Computer Vision, {ICCV}
  2017, Venice, Italy, October 22-29, 2017}, 2017.

\end{thebibliography}
}

\clearpage
\setcounter{section}{0}
\begin{center}
    {\Large \textsc{Supplementary Material}}
\end{center}

\section{Generated Data Influence on Soiling Semantic Segmentation}

\begin{figure*}[tb]
    \centering
    \includegraphics[width=\linewidth]{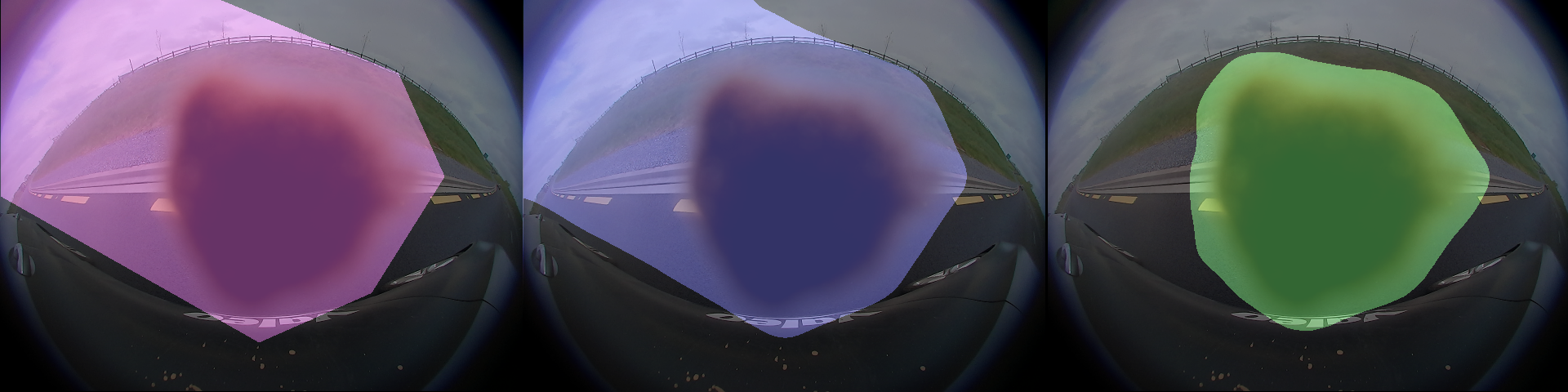}
    \caption{The problem of weak annotation labels and how generated data can help, presented on a testing image, which was not used during the training. The images come from the \emph{WoodScape Dataset}. Left: original weak (polygonal) annotation; Note, that not only it is very coarse, but also prone to errors. Middle: semantic segmentation output, when trained on the weak annotation labels in a fully supervised manner. Right: the same semantic segmentation network output, however, in this case also the generated data were used. Note, how having precise annotation labels for the generated data help in refining the segmentation output.}
    \label{fig:app:SegWeakAnnIssues}
\end{figure*}

In Figure~\ref{fig:app:SegWeakAnnIssues}, we depict the problems, we are facing when using the \emph{WoodScape Dataset}. Because only coarse polygonal annotation is available (which is moreover prone to errors), the classical fully supervised training of the semantic segmentation for the soiling is affected by the annotation quality. If we, however, add the generated images along with the precisely generated annotations, the same fully supervised training exploits these annotations, with a desired effect on the semantic segmentation output quality. Thus, in addition to providing novel patterns for augmentation in training, the proposed method also generates precise annotation compared to coarse manual annotation.

\section{Network Architectures}

In this section, we provide details about the network architectures we used in our experiments. 

\subsection{Baseline Networks Architecture}

\begin{figure}[tbh!]
    \centering
    \includegraphics[width=0.5\linewidth]{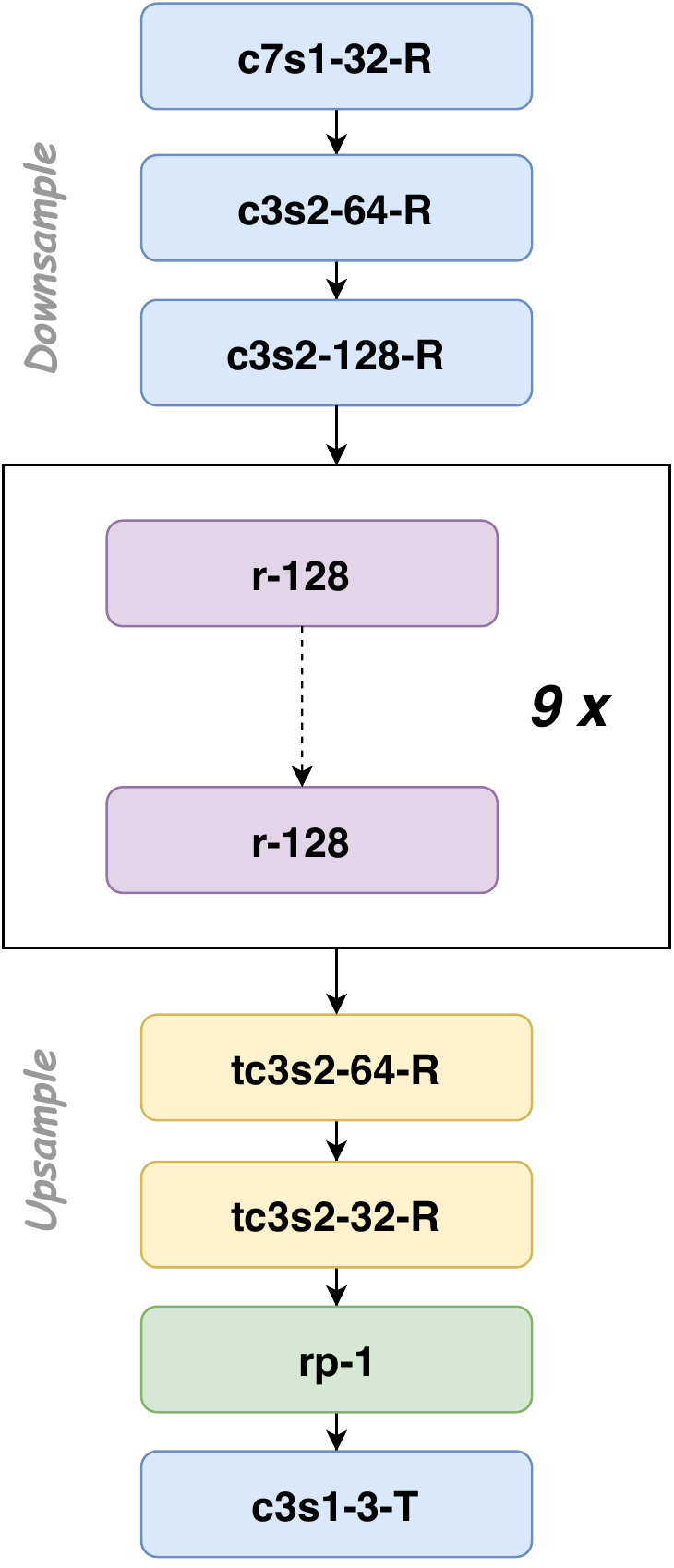}
    \caption{The Generator's architecture, used in the CycleGAN. ``c7s1-32-R'' reads as $2$D convolutional layer with kernel size $7\times7$ pixels, stride $1$, $32$ output channels, followed by a ReLU activation layer. $T$ stands for the $\mathrm{tanh}$ activation; ``tc'' is a shortcut for the transposed convolution; ``rp'' means reflection padding; ``r-128'' is a shorthand for residual block with $128$ channels.}
    \label{fig:app:generator}
\end{figure}

\begin{figure}[tbh!]
    \centering
    \includegraphics[width=0.3\linewidth]{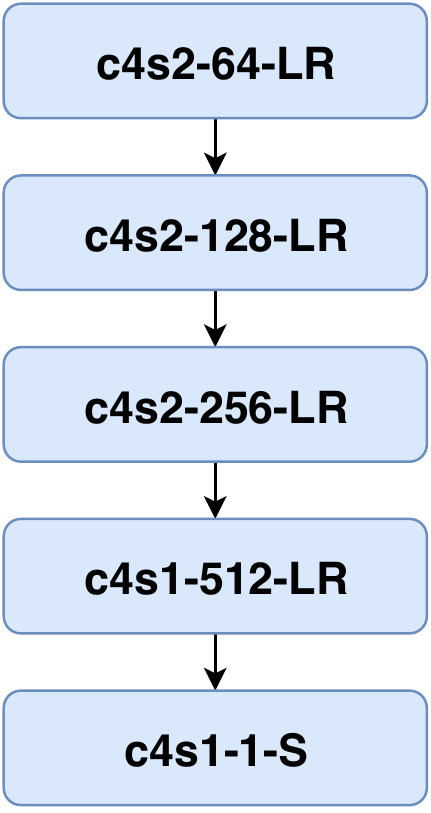}
    \caption{The Discriminator's architecture, used in the CycleGAN. ``c4s2-64-LR'' reads as $2$D convolutional layer with kernel size $4\times4$ pixels, stride $2$, $64$ output channels, followed by a Leaky ReLU activation layer. $S$ stands for the sigmoid activation layer.}
    \label{fig:app:discriminator}
\end{figure}

\begin{figure}[tbh!]
    \centering
    \includegraphics[width=0.45\linewidth]{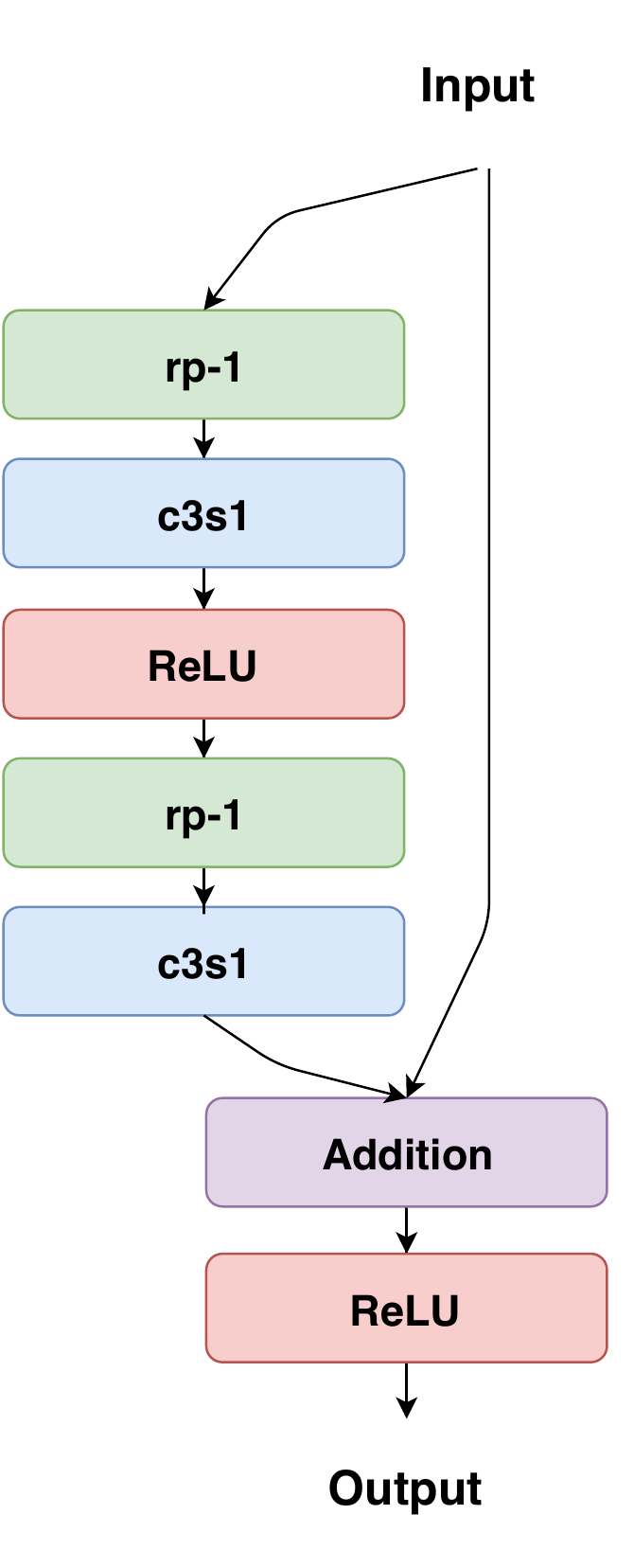}
    \caption{The residual block, which is used in the Generator's architecture. Note, that convolutional blocks ``c3s1'' might be followed by a BatchNorm layer (used only in the soiling mask segmentation network).}
    \label{fig:app:resblock}
\end{figure}

\paragraph{Generator and Discriminator of CycleGAN} In the CycleGAN generator network (see Figure~\ref{fig:app:generator}), We use the residual block depicted in Figure~\ref{fig:app:resblock}. Note, that we do not use any kind of normalization, like BatchNorm or InstanceNorm. In Discriminator (see Figure~\ref{fig:app:discriminator}, we use Leaky ReLU and a fully convolutional classification layer at the end. 

\begin{figure}[tbh!]
    \centering
    \includegraphics[width=0.3\linewidth]{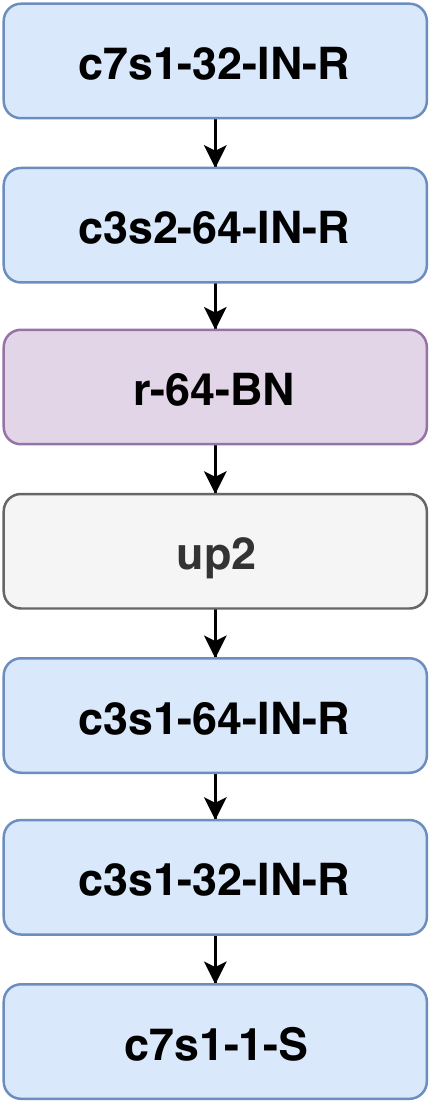}
    \caption{The soiling mask segmentation network. The same type of shorthands as in previous figures are used with an addition of ``IN'' for Instance Normalization and ``BN'' for BatchNorm layers. ``up2'' is a simple nearest neighbor upsampling with scale factor $2$.}
    \label{fig:app:mask_seg}
\end{figure}

\paragraph{Soiling Mask Segmentation} In Figure~\ref{fig:app:mask_seg}, we show the semantic segmentation network for the soiling mask detection. For the soiling mask segmentation network, we use Instance Normalization in convolutional blocks and BatchNorm in the residual block. 

\begin{figure}[tbh!]
    \centering
    \includegraphics[width=0.65\linewidth]{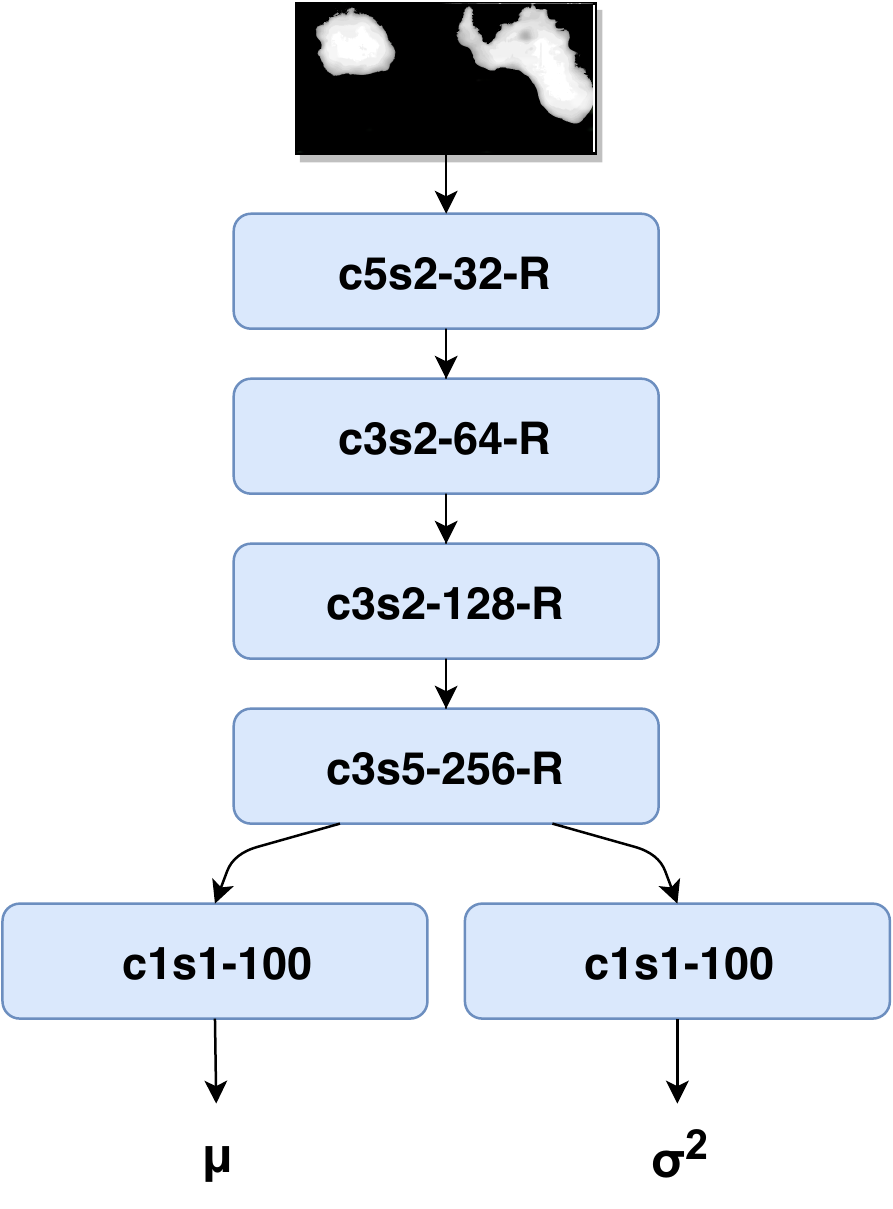}
    \caption{The architecture of the VAE encoder used in our experiments. The notation ``c3s2-64-R'' corresponds to a $2$D convolutional layer, with kernel size $3 \times 3$ pixels, stride of $2$, $64$ output channels, followed by a ReLU activation layer.}
    \label{fig:app:VAE_enc}
\end{figure}

\begin{figure}[tbh!]
    \centering
    \includegraphics[width=0.3\linewidth]{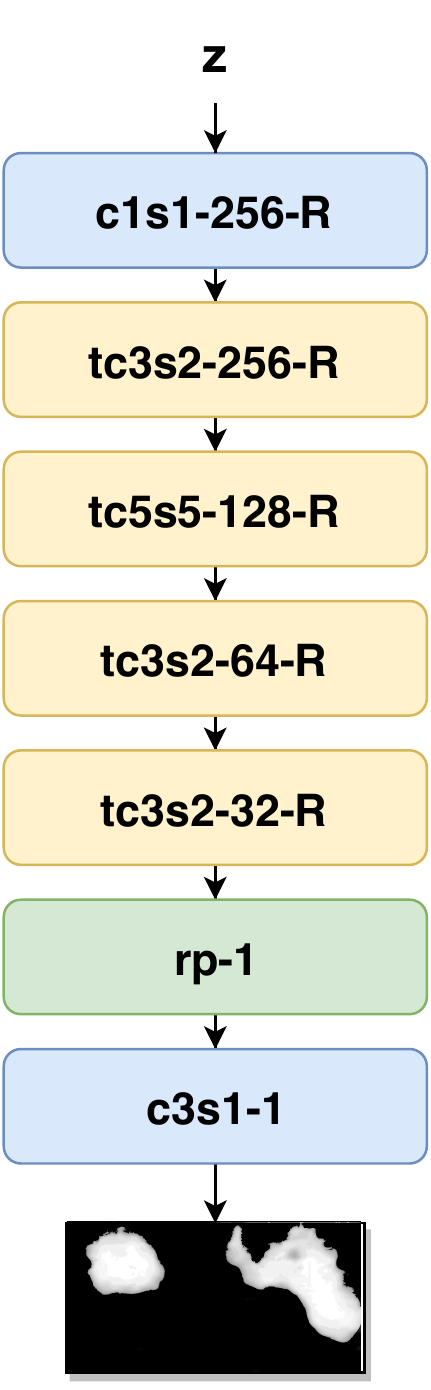}
    \caption{The architecture of the VAE decoder used in our experiments. The notation ``c3s2-64-R'' corresponds to a $2$D convolutional layer, with kernel size $3 \times 3$ pixels, stride of $2$, $64$ output channels, followed by a ReLU activation layer. ``tc3s2-256-R'' represents a transposed convolution, with kernel $3 \times 3$ pixels, stride $2$, $256$ output channels, followed by ReLU activation. ``rp-1'' means a reflection padding of size $1$.}
    \label{fig:app:VAE_dec}
\end{figure}

\paragraph{Variational AutoEncoder} We depict the encoder part of the VAE in Figure~\ref{fig:app:VAE_enc}. The $\#\mu$ and $\#\sigma$ are combined into $\#z$ via reparametrization trick. The VAE's decoder is shown in Figure~\ref{fig:app:VAE_dec}. 

\subsection{DirtyGAN Architecture}

\begin{figure*}[tb]
    \centering
    \includegraphics[width=\textwidth]{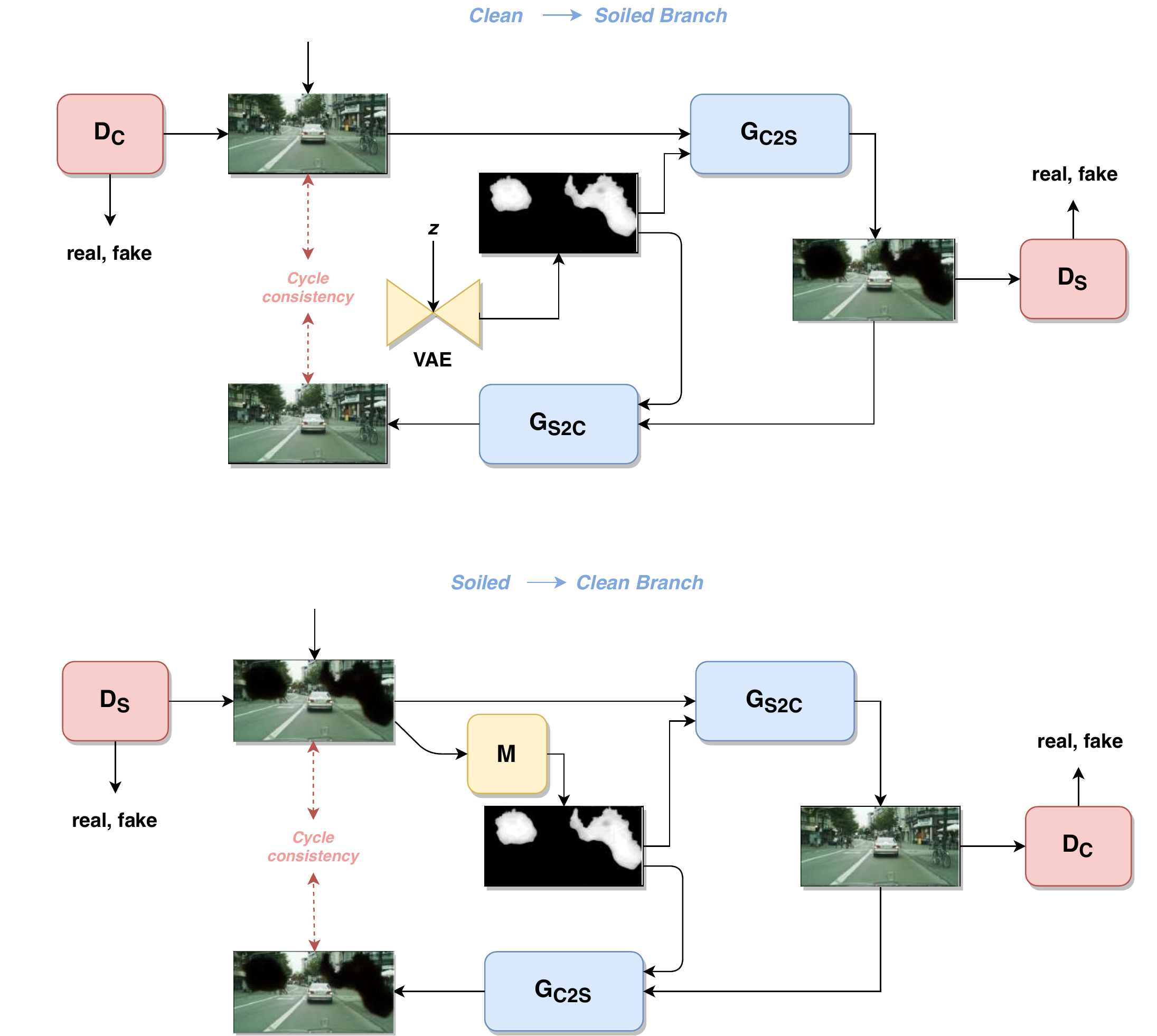}
    \caption{DirtyGAN scheme.}
    \label{fig:DirtyGAN}
\end{figure*}

DirtyGAN is composed of the building blocks used in the baseline algorithm, therefore the networks architecture is mostly the same, only with some minor tweaks. The DirtyGAN scheme is depicted in Figure~\ref{fig:DirtyGAN}.

\section{Generated Data}

\begin{figure*}
    \centering
    \includegraphics[width=0.9\linewidth]{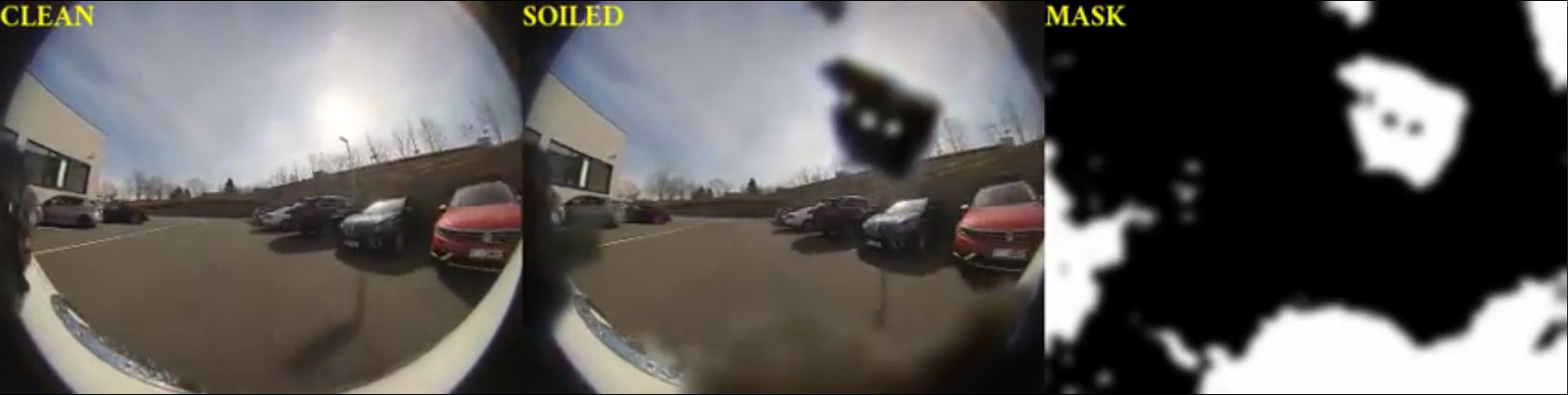}
    \includegraphics[width=0.9\linewidth]{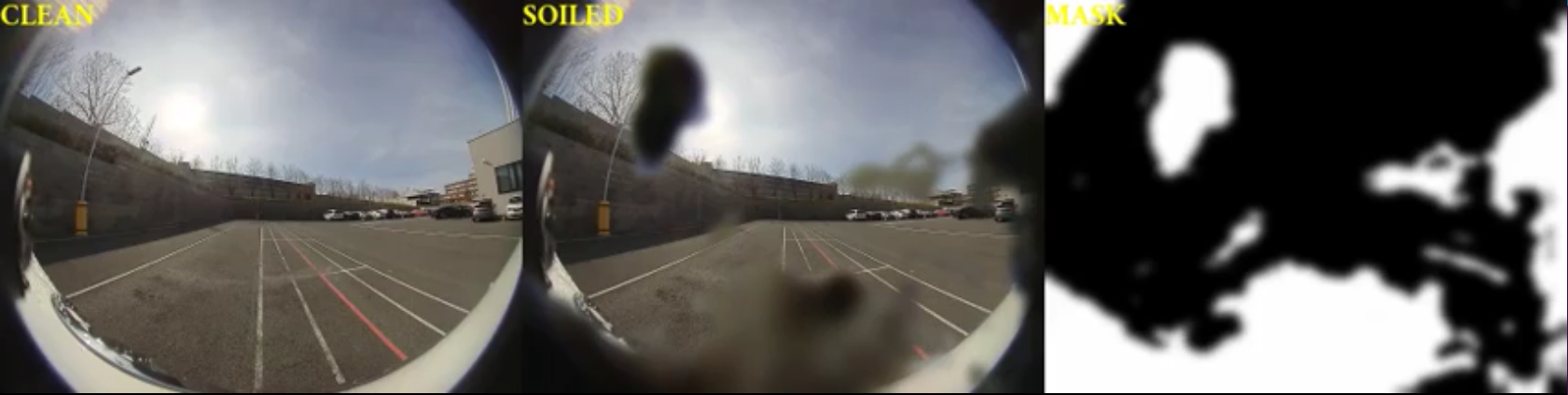}
    \includegraphics[width=0.9\linewidth]{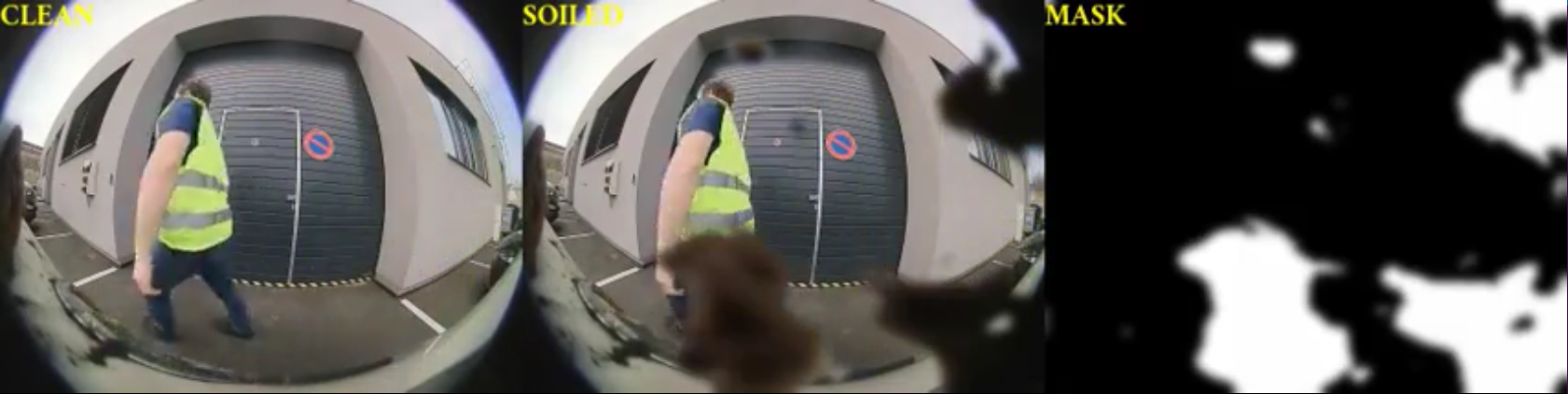}
    \includegraphics[width=0.9\linewidth]{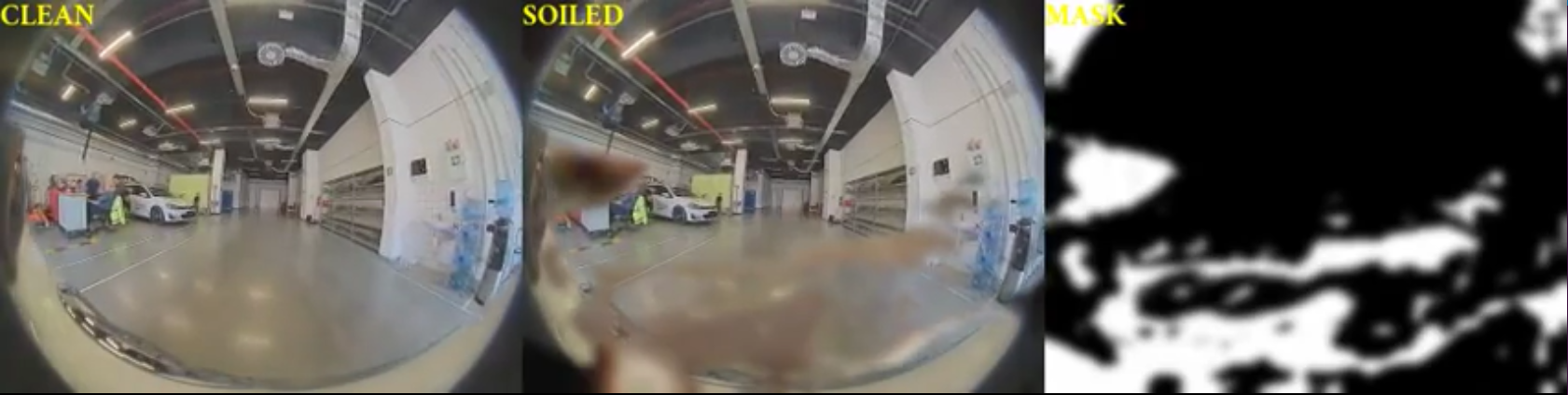}
    \includegraphics[width=0.9\linewidth]{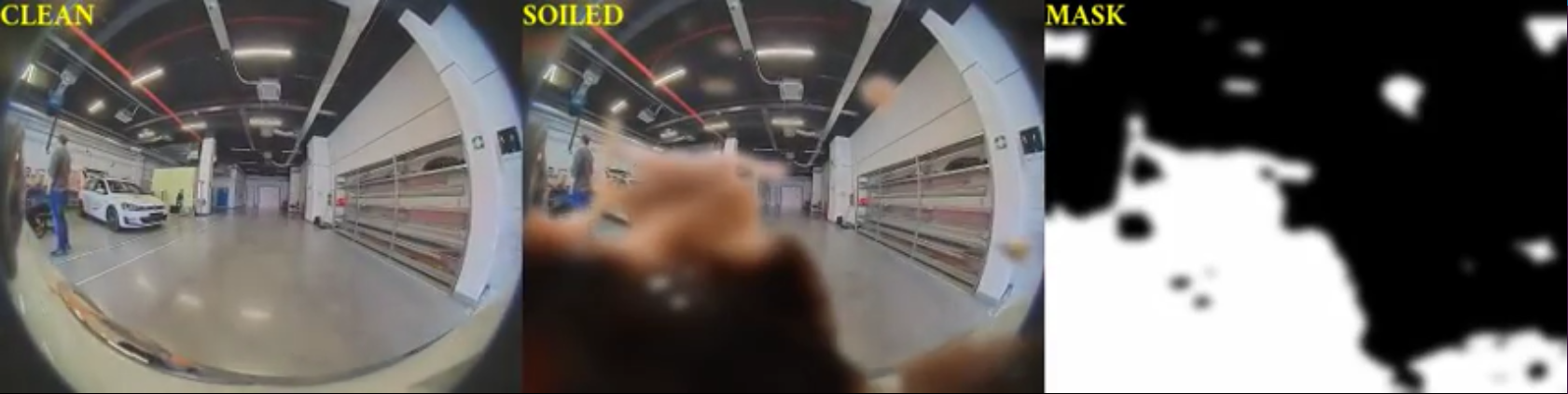}
    \caption{Several frames extracted from the supplementary video files. Please, see the original video files for the best experience.}
    \label{fig:app:frames}
\end{figure*}

\begin{figure*}
    \centering
    \includegraphics[width=0.9\linewidth]{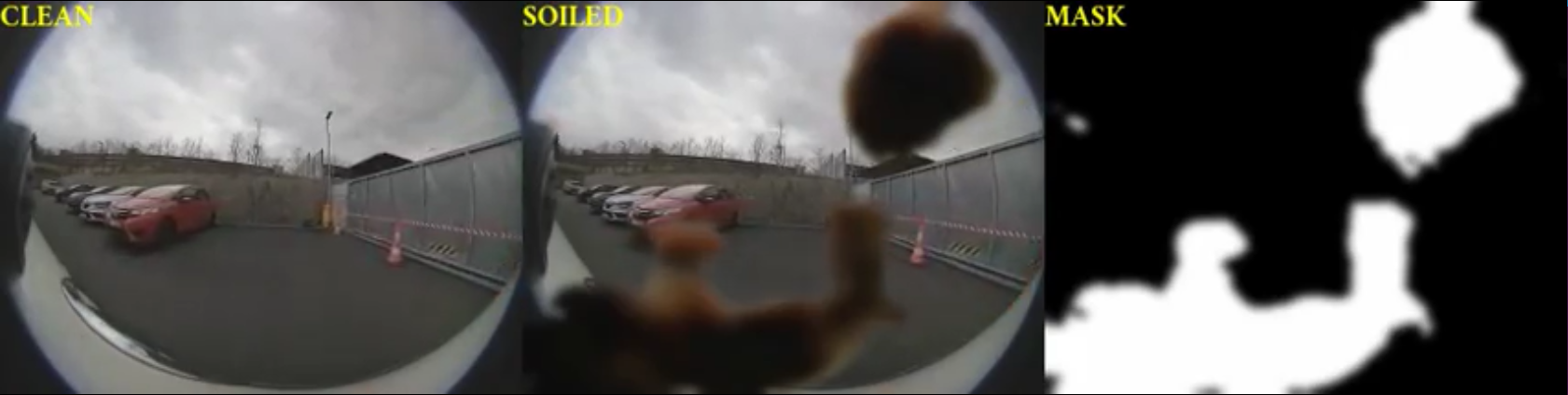}
    \includegraphics[width=0.9\linewidth]{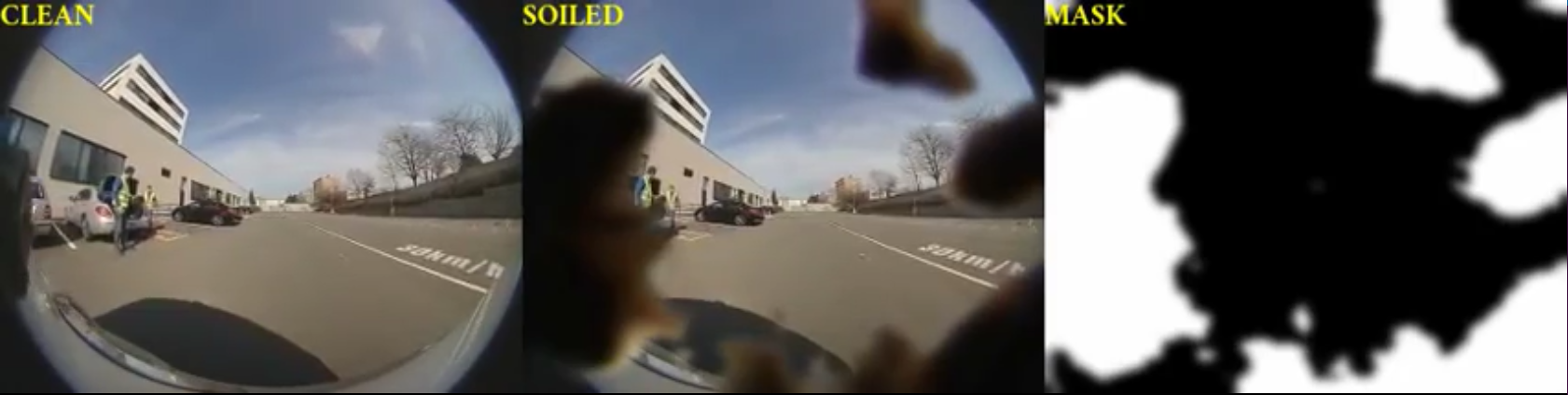}
    \includegraphics[width=0.9\linewidth]{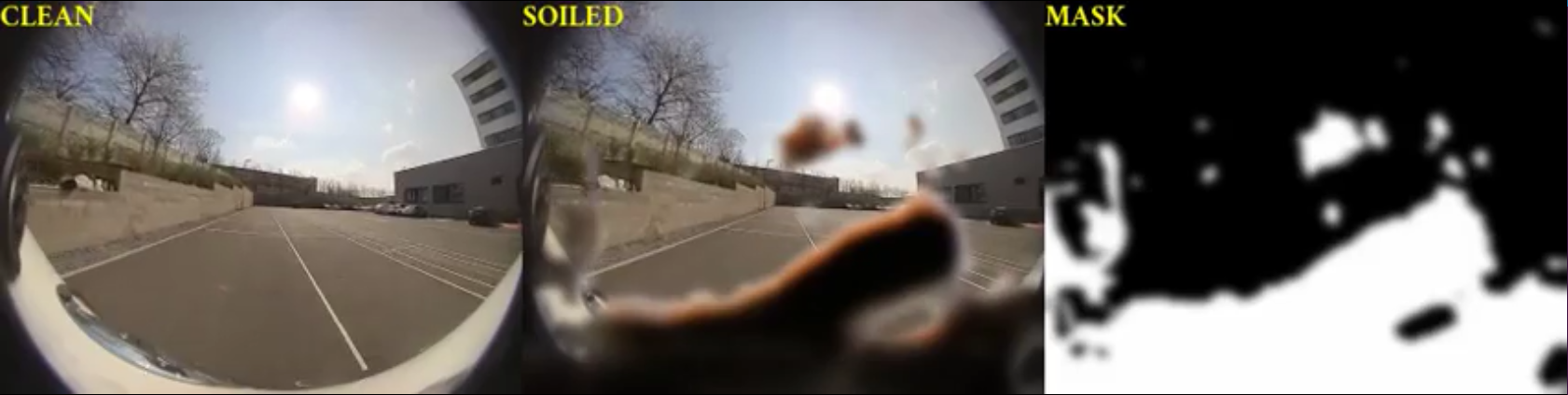}
    \includegraphics[width=0.9\linewidth]{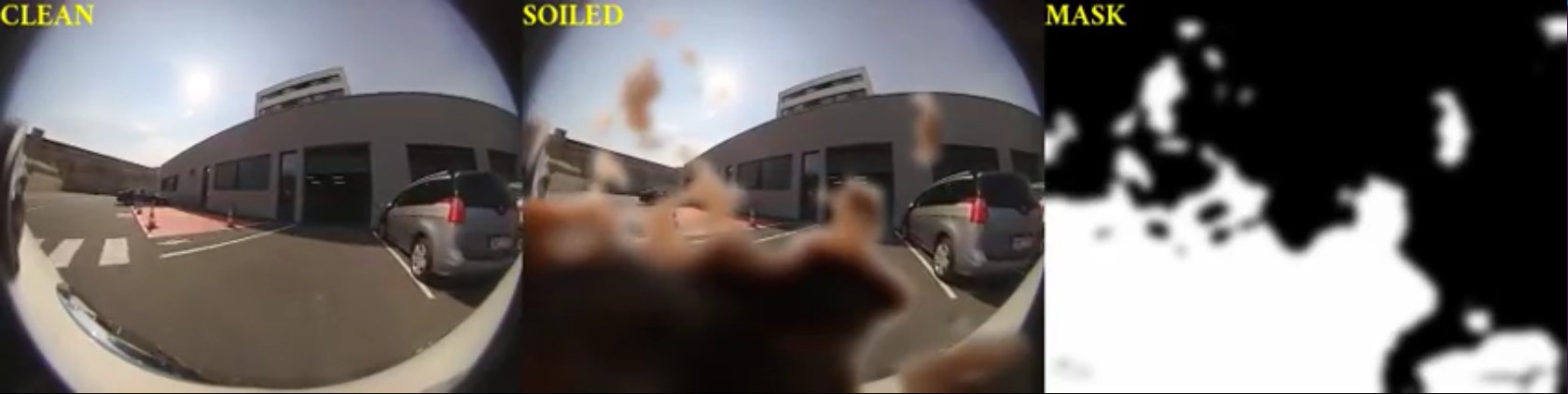}
    \includegraphics[width=0.9\linewidth]{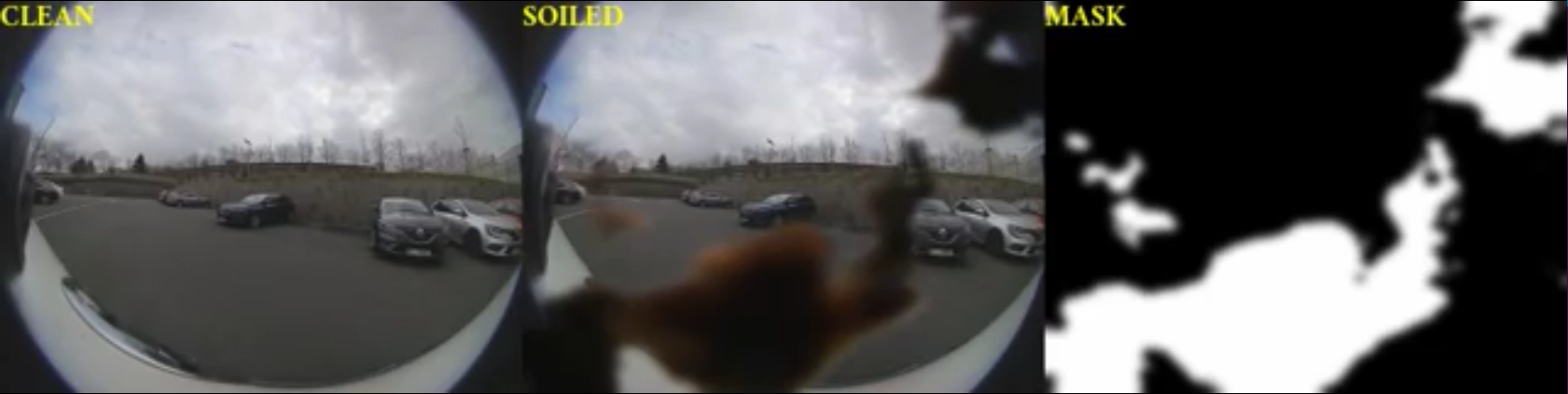}
    \caption{Several frames extracted from the supplementary video files. Please, see the original video files for the best experience.}
    \label{fig:app:frames2}
\end{figure*}

In Figure~\ref{fig:app:frames} and \ref{fig:app:frames2}, we depict random frame from generated videos, which we enclose to the supplementary. To fit the size limit, we resized the images by $1/4$ for both width and height. The first image in the row is the original frame from the video sequence as it was recorded. The second image in the row is the artificially generated soiled version of the first image and the third image in the row is the soiling mask, which can serve as the automatic annotation for the soiled image. Although it is possible to use our algorithm to generate dynamic soiling effects, we present here only the static soiling. Please, see the video files for the best experience. We have shared a subset of 11 videos with different soiling patterns.

\end{document}